\documentclass[journal,twoside,web]{ieeecolor}
\usepackage{jsen}
\usepackage{cite}
\usepackage{amsmath,amssymb,amsfonts}
\usepackage{graphicx}
\usepackage{textcomp}
\usepackage{wrapfig}
\usepackage{url}
\usepackage{multirow}
\usepackage{diagbox} 

\def\BibTeX{{\rm B\kern-.05em{\sc i\kern-.025em b}\kern-.08em
    T\kern-.1667em\lower.7ex\hbox{E}\kern-.125emX}}
\definecolor{abstractbg}{rgb}{0.89804,0.94510,0.83137}
\usepackage{cite}
\usepackage{booktabs} 
\usepackage{ifpdf}
\usepackage{url}
\usepackage{xspace}
\usepackage{balance}
\usepackage{caption}
\usepackage[tight,footnotesize]{subfigure}
\usepackage{multirow}
\usepackage{color}
\usepackage{comment}
\usepackage{algorithm}
\usepackage{algpseudocode}
\usepackage{tabularx}

\floatname{algorithm}{Algorithm}

\usepackage{makecell} 

\usepackage{mathtools}

\newcommand{\etal}{\textit{et al}.\xspace}
\newcommand{\ie}{\textit{i}.\textit{e}.\xspace}

\newcommand{\revised}[1]{\textcolor{black}{#1}}

\begin{document}

\title{Real-Time Fall Detection Using Smartphone Accelerometers and WiFi Channel State Information}
\author{         
Lingyun~Wang,     
        Deqi~Su,
        Aohua~Zhang,
	Yujun~Zhu,
	Weiwei~Jiang,
Xin~He,~\IEEEmembership{Member,~IEEE},
Panlong~Yang,~\IEEEmembership{Senior Member,~IEEE}
\thanks{This work has been in part supported by the Natural Science Foundation of China under grant No. 62072004. Corresponding author: {\it Xin He, Yujun Zhu.}}
\thanks{X. He, D. Su, L. Wang, A. Zhang, and Y. Zhu are with the School of Computer and Information, Anhui Normal University, Wuhu, 241002, Anhui, China (e-mail: \{lingyunwang, zhangaohua, sudeqi, xin.he, zhuyujun\}@ahnu.edu.cn).}
\thanks{W. Jiang and P. Yang are with the School of Computer Science, Nanjing University of Information Science and Technology, 210044, Jiangsu, China (e-mail: \{weiwei.jiang, plyang\}@nuist.edu.cn). }
}

\IEEEtitleabstractindextext{%
\fcolorbox{abstractbg}{abstractbg}{%
\begin{minipage}{\textwidth}%
\begin{wrapfigure}[12]{r}{3in}%
\includegraphics[width=2.8in]{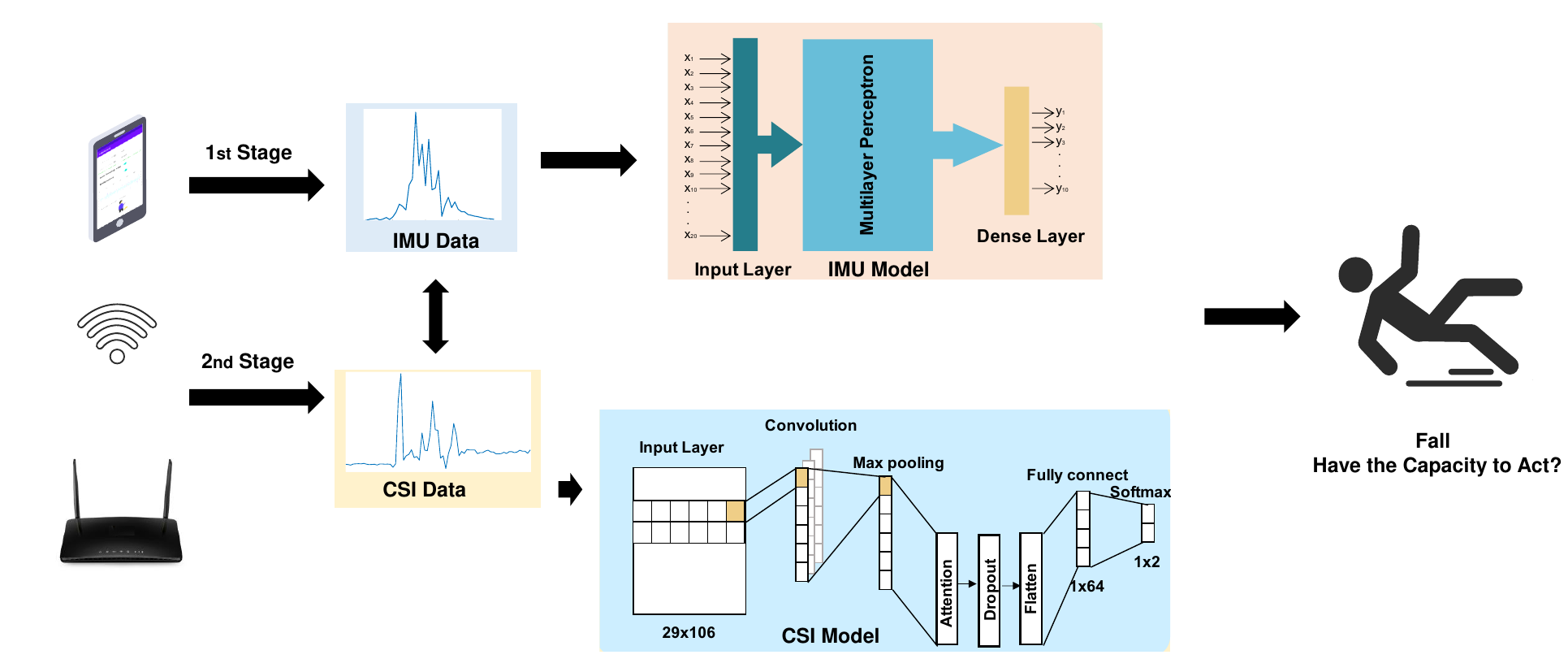}%
\end{wrapfigure}%
\begin{abstract}
In recent years, as the population ages, falls have increasingly posed a significant threat to the health of the elderly. We propose a real-time fall detection system that integrates the inertial measurement unit (IMU) of a smartphone with optimized Wi-Fi channel state information (CSI) for secondary validation. Initially, the IMU distinguishes falls from routine daily activities with minimal computational demand. Subsequently, the CSI is employed for further assessment, which includes evaluating the individual's post-fall mobility. This methodology not only achieves high accuracy but also reduces energy consumption in the smartphone platform. An Android application developed specifically for the purpose issues an emergency alert if the user experiences a fall and is unable to move. Experimental results indicate that the CSI model, based on convolutional neural networks (CNN), achieves a detection accuracy of 99\%, \revised{surpassing comparable IMU-only models, and demonstrating significant resilience in distinguishing between falls and non-fall activities.}
\end{abstract}

\end{minipage}}}

\maketitle

\section{Introduction}
Falls constitute the second most common cause of unintentional injury-related fatalities and are a significant factor that causes injuries and deaths among older adults. Such incidents commonly result in fractures, head injuries, and other grave health issues. According to estimated data collected by WHO~\cite{Falls2021}, each year falls result in 684,000 fatalities, predominantly in low- and middle-income countries. Annually, 37.3 million falls are severe enough to require medical attention, leading to significant health impacts, long-term care needs, and substantial financial costs~\cite{thakur2021country,koda2018relationship,khanmohammadi2017time,ludwig2012health}.

Fall detection hence is of significant importance. Researches carried out in this domain encompass a broad range of technologies and methodologies aimed at identifying falls, particularly among the elderly~\cite{hu2022}, to prevent serious injuries and fatalities. The primary approaches include wearable sensors, ambient sensors, and vision-based systems~\cite{ren2019research,madansingh2015smartphone,tolkiehn2011direction,basu2023advancements,nakamura2020wi,chowdhury2018using,ramezani2018sensing}. 
Wearable sensors, such as accelerometers and gyroscopes, are widely employed to detect falls through the analysis of motion patterns. \revised{They offer high accuracy and function effectively across diverse environments. However, their efficacy is contingent upon user compliance, which poses a significant challenge, particularly among elderly users who may forget or neglect to wear these devices~\cite{lara2012survey}.}
\revised{In contrast,} ambient sensors, including infrared and thermal sensors, monitor environmental changes to identify falls without \revised{user involvement, effectively preserving privacy. However, their effectiveness is limited to areas within sensor range, making extensive coverage potentially costly~\cite{Zhang2016feasibility}.} 
Vision-based systems, \revised{which} utilize cameras and computer vision techniques to detect falls by analyzing visual data, \revised{can achieve highly precise fall detection autonomously, yet they pose significant privacy concerns and are generally fixed in specific locations, requiring extensive setup~\cite{gaya2024deep}.} 
Recent advancements in fall detection have focused on fusion-based methods, which combine data from multiple sensors to improve accuracy and reliability, \revised{thus enhancing overall reliability~\cite{Cagnoni2009sensor}.} This trend is evident in the integration of radio frequency (RF) signals with traditional sensor data to enhance detection capabilities\revised{~\cite{krovvidi2024activity}}. 

For instance, Ponce~\cite{ponce2020sensor}~\etal demonstrate the effectiveness of combining wearable sensors and vision devices to achieve high accuracy in fall detection by analyzing sensor locations and minimal deployment strategies. Sowmya and Pillai~\cite{sowmya2021human} highlight the potential of using machine learning algorithms with wearable sensors to detect falls, showing that ensemble techniques like random forest classifiers can achieve up to 98\% accuracy. Additionally, Ramanujam and Padmavathi~\cite {ramanujam2019vision} offer non-invasive and cost-effective solutions, with their research indicating that these systems can outperform traditional sensor-based methods in activity recognition and fall detection. Khan~\cite{khan2023contactless}~\etal came up with the idea of integrating RF signals with traditional methods to further enhance detection capabilities, addressing privacy concerns associated with wearable and vision-based technologies. These advancements highlight the ongoing efforts to improve fall detection systems through the integration of diverse sensor technologies and sophisticated algorithms.

Researchers in fall detection face several significant challenges. A major challenge in fall detection is the high rate of false positives, where non-fall activities such as sitting down quickly or abrupt movements are misinterpreted as falls. Lo~\cite{lo2021fpga}~\etal emphasize the critical need for sophisticated algorithms in their FPGA-based study that can more effectively differentiate between fall and non-fall events, including situations where individuals lie down quickly or end in a sitting position. This necessity underscores the importance of developing advanced detection algorithms to minimize false positives and enhance the reliability of fall detection systems. 

Real-time processing of fall detection data is critical for timely intervention but presents a significant technological challenge due to the computational demands involved. The use of millimeter wave signals in systems like mmFall~\cite{Li2022} exemplifies a solution that achieves high accuracy while maintaining low computational complexity which utilizes spatial-temporal processing alongside a lightweight convolutional neural network (CNN). However, the high cost of such advanced systems remains a considerable barrier to widespread adoption. 

User compliance is another problem, as the effectiveness of wearable sensors depends on consistent use by the elderly, who may forget or refuse to wear the devices~\cite{baig2019systematic}. Furthermore, there is a lack of extensive real-world datasets that include actual falls by elderly individuals, which limits the ability to validate and improve detection algorithms effectively~\cite{naranjo2012personalization,denkovski2022multi}. Ethical concerns also prevent the collection of such data, leading to reliance on simulated falls by younger individuals, which may not accurately represent real fall dynamics. 

\revised{Unlike existing fall detection methods that primarily rely on a single sensor modality, this study introduces a cross-modal fusion strategy for real-time fall verification.}
\revised{Our system} combines data from both the inertial measurement unit (IMU) and channel state information (CSI) via smartphones. The system initiates a data collection phase, during which IMU and CSI data from various fall scenarios are collected using smartphones. The collected data undergoes preprocessing to eliminate noise and normalize the inputs, which ensures the quality and consistency of the data. The refined data is subsequently inputted into two distinct models: a multi-layer perceptron (MLP) model processes the IMU data, and a CNN model handles the CSI data. These models are trained to distinguish falls by identifying the unique characteristics of fall events as opposed to normal activities. The outputs from these models are integrated to improve detection accuracy and minimize false positives, thereby providing a reliable solution for real-time fall detection.

Several innovative contributions of our work to the field of fall detection can be summarized as follows:
\begin{itemize} \item[1)] We propose a real-time fall detection system that only requires a smartphone connected to home Wi-Fi, eliminating the need for additional tools. Our developed app implements AI algorithms for fall detection, offering a convenient, low-power solution easily integrated into daily life, especially for the elderly or those at higher risk of falls.

\item[2)] Our approach combines sensor technology (accelerometers and gyroscopes) with wireless communication perception. The system uses a two-stage detection process: first, detecting falls via IMU sensors, and then double-checking the user's ability to move post-fall. Additionally, the CSI-based detection ensures continuous monitoring within indoor spaces, even without the user carrying a device, enhancing safety in settings like elderly care facilities. \end{itemize}

We yield several important findings and concrete results. By integrating IMU and CSI data, we significantly reduced the number of false positives and improved the overall accuracy of fall detection. Our system is able to accurately distinguish between falls and other rapid movements, such as quickly picking up a phone, demonstrating its practical applicability. 

The real-world testing of our system shows high accuracy rates in detecting falls, validating the effectiveness of our approach. Furthermore, the use of CSI data as a secondary validation step further enhanced the system's reliability, providing an extra layer of assurance against false positives. The overall recognition accuracy of our system is 99\%, which highlights the potential of our integrated approach to offer a reliable and efficient solution for fall detection, with the capability to be deployed in real-world scenarios. 

\begin{figure}[t]
    \centering
    \includegraphics[width=3in]{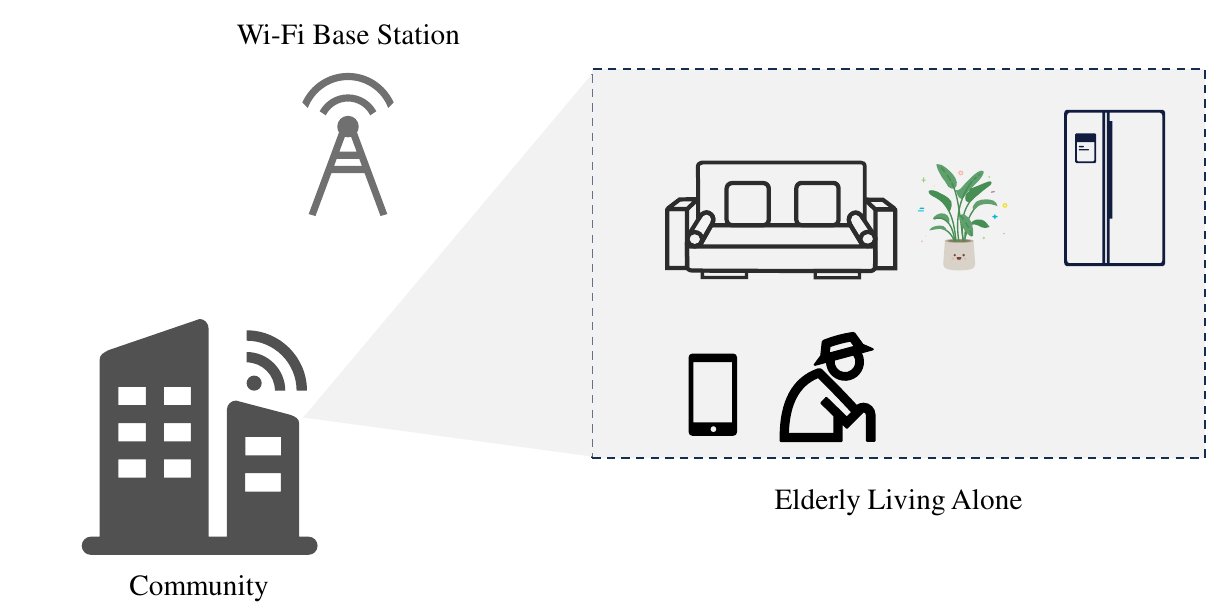}
    \caption{\revised{An example application scenario for fall detection using Wi-Fi CSI is monitoring elderly individuals living alone, where our system can immediately detect a fall and issue warnings when necessary.}}
    \label{fig:sys_model}
\end{figure}

The rest of this paper is organized as follows: Section II details the system design of the proposed fall detection system, including data acquisition, data preprocessing, feature extraction, and the learning model design. Section III examines the performance of our designed system from multiple perspectives, comparing it with other models. Section IV and Section V conclude and forecast the research direction.

\begin{figure*}[t]
    \centering
    \includegraphics[width=0.95\linewidth]{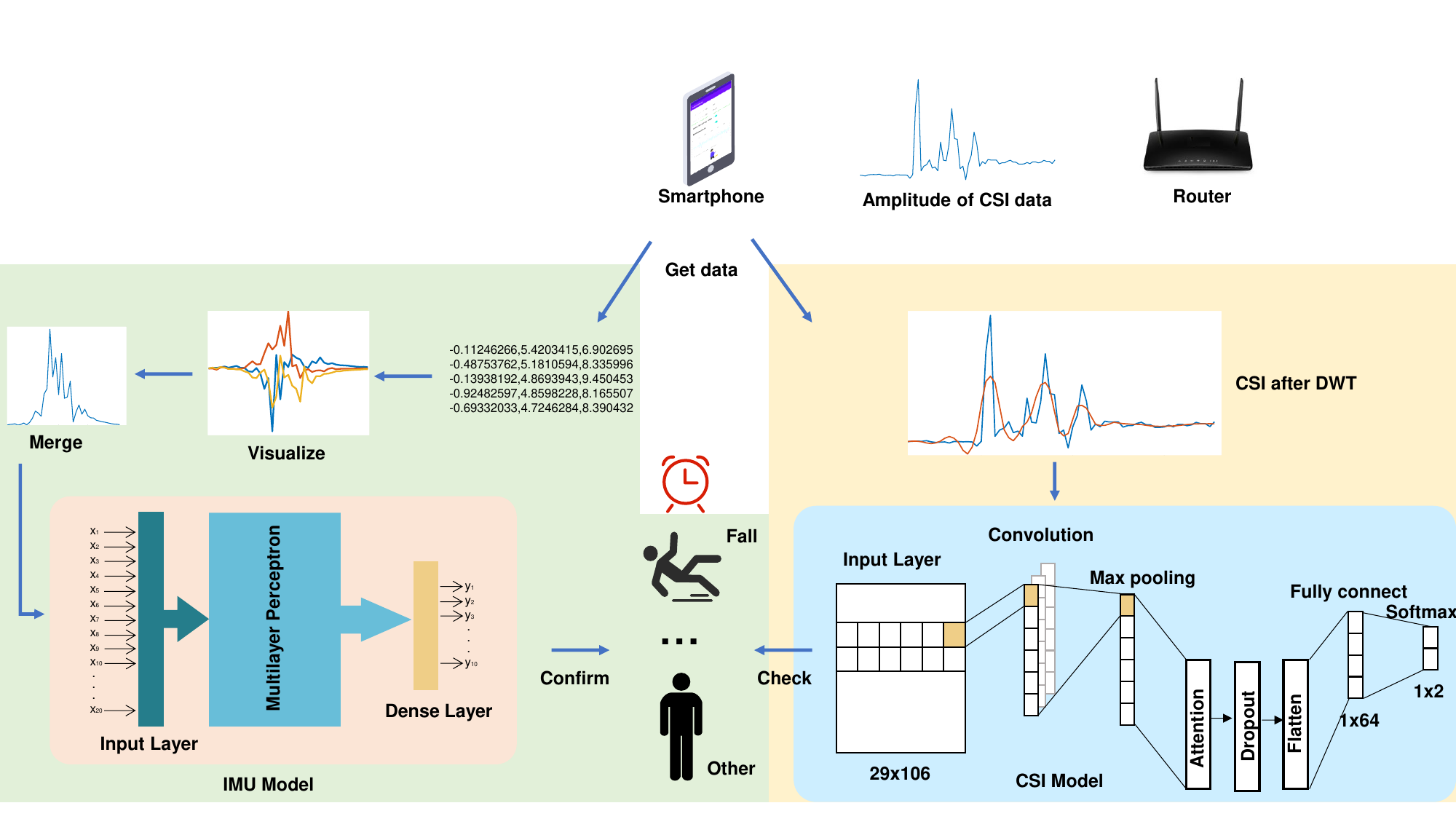}
    \caption{The framework of the real-time fall detection system.}
    \label{fig:sys_frame}
\end{figure*}
\section{System Design}
This section outlines our system configuration utilizing an 802.11 wireless network interface card (NIC) to capture CSI data. The employed smartphone is the OPPO Reno9 Pro, which features a MediaTek 8100-MAX CPU and a battery capacity of 4500~mAh, while the computer used for data processing and model training is outfitted with an NVIDIA GeForce RTX 3060 GPU.

\subsection{Overview of the system}
We design a smartphone-based system for real-time fall detection and alert generation that integrates data from the built-in accelerometer and gyroscope, alongside CSI, to identify and respond to falls.
\revised{As depicted in Fig.~\ref{fig:sys_model}, our system is primarily designed for one application scenario in which elderly individuals living independently.}

Initially, raw data from the accelerometer and gyroscope are collected and subjected to normalization to ensure consistency in the data scale and facilitate accurate analysis. 
Subsequent feature extraction processes identify characteristics specific to fall patterns. These features are then used in a classification algorithm to determine the occurrence of a fall. 
Concurrently, the system processes CSI data by first calculating the amplitude of the Wi-Fi signals to gauge environmental interactions, followed by computing the rate of change in this amplitude to assess the dynamics and severity of the movement. 

Our system further evaluates the individual's capacity to act post-fall.
If the person can move, the system issues a reminder indicating that although a fall has been detected, the individual remains active and does not require assistance. Oppositively, 
if the individual is incapacitated, a warning alert that the person has fallen and is unable to move, triggers an emergency response.
This dual-analysis approach not only detects falls but also assesses the condition of the individual afterward, enabling timely and appropriate interventions based on the severity of the incident and the person's immediate needs. The framework of the fall detection system is shown in 
Fig.~\ref{fig:sys_frame}.


\subsection{System setups}
\begin{figure*}[t]
\centering
\begin{tabular}{ccc}
\includegraphics[width=0.33\textwidth]{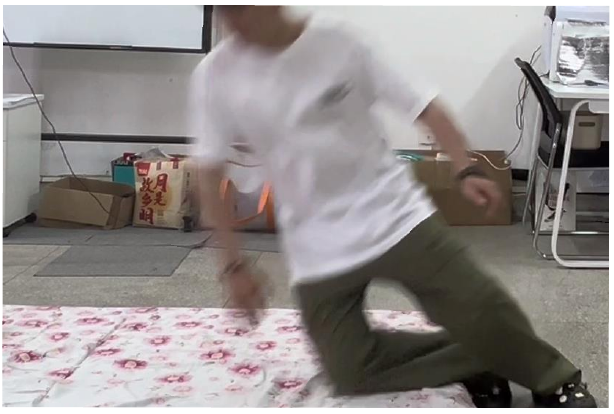} &
\includegraphics[width=0.33\textwidth]{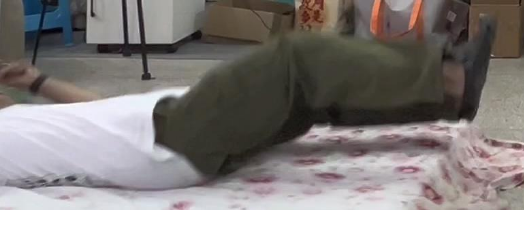} &
\includegraphics[width=0.33\textwidth]{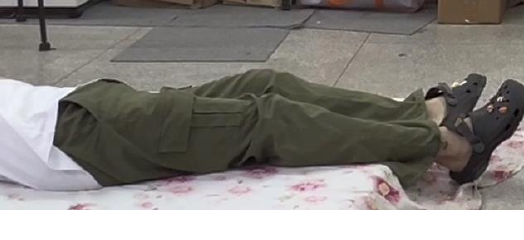} \\
(a) descent phase & (b) impact phase & (c) stationary phase 
\end{tabular}
\caption{Three phases of a fall event: descent, impact, and stationary.}
\label{fig:three_phase}
\end{figure*}

In daily activities, human motion is generally smooth and continuous, with actions such as walking, sitting, and turning to exhibit regularity and stability. However, during a fall, the body's movements display sudden, unstable characteristics such as rapid descent, loss of balance, and collapse. These abrupt changes in motion patterns contrast sharply with everyday behaviors and are crucial features for identifying fall-related actions. \revised{We divide the falling process into three phases: descent phase, impact phase, and stationary phase}, as shown in Fig.~\ref{fig:three_phase}. 

\textbf{Descent phase}.
During the descent phase of a fall, the body experiences rapid changes in acceleration as it transitions from any potential pre-fall state to a rapid downward fall. In addition to \revised{this}, the body often undergoes rotational and tilting motions, especially after losing balance. These movements are attempts to maintain posture or find a support point. Compared to normal walking or other activities, this phase typically exhibits irregular motion patterns, characterized by more sudden movements. Fig.~\ref{fig:three_phase}(a) illustrates the descent phase of a fall. 

\textbf{Impact phase}.
As shown in Fig.~\ref{fig:three_phase}(b), during the impact phase, when a person falls and contacts the ground, a collision occurs between the body and the surface, which generates stress and propagates shock waves. At this moment, the change in acceleration is extremely abrupt. After hitting the ground, a rebound might occur. This rebound is due to some of the energy absorbed during the fall being released in some form after the collision, causing the body to momentarily lift off the ground again. This phase is critical for detecting falls, as the acceleration patterns observed differ markedly from those observed during normal activities or less severe falls.

\textbf{Stationary phase}.
After a fall occurs, it can be categorized into two types: falls from which one can recover independently, and falls from which one cannot. We primarily focus on non-recoverable falls, which are characterized by the individual lying on the ground, unable to move or seek help on their own. It is only in these situations that a fall is finally classified as a non-recoverable fall, which is the specific type of fall behavior that our research aims to monitor. The stationary phase is shown in Fig.~\ref{fig:three_phase}(c).

We collected data on ten different types of actions. Table~\ref{tab:actions} lists the types of actions and the number of data samples collected for each. The goal is to accurately detect falls from among these various actions.
\begin{table}[t]
    \centering
    \caption{\revised{Composition of the collected dataset.}}
    \begin{tabular}{c|c|c}
        \hline
        \textbf{\revised{Activity}} & \textbf{\revised{Explanation}} & \textbf{\revised{Data Size}} \\ \hline
        \revised{fall} & \revised{fall down} & \revised{820} \\ 
        \revised{play} & \revised{play with a smartphone} & \revised{748} \\
        \revised{pickput} & \revised{pick up and put down a phone} & \revised{870} \\
        \revised{sit} & \revised{sit down and stand up} & \revised{876} \\ 
        \revised{squat} & \revised{squat down and stand up} & \revised{818} \\ 
        \revised{static} & \revised{stay still} & \revised{896} \\ 
        \revised{walk} & \revised{walk around} & \revised{759} \\ 
        \revised{up\_down} & \revised{put down and pick up} & \revised{832} \\ 
        \revised{toward\_down} & \revised{swing a phone quickly} & \revised{866} \\ 
        \revised{throwing} & \revised{throwing a phone} & \revised{820} \\ \hline
    \end{tabular}
    \label{tab:actions}
\end{table}

These ten types of movements can be divided into three main categories. The first category is falls. The second category consists of actions commonly seen indoors in daily life, including play, \ie, playing with a smartphone, pick and put, \ie, picking up and putting down a phone, sit, \ie, sitting down and standing up, squat, \ie, squatting down and standing up, static, and walk. The third category comprises some less common actions but, when analyzed through accelerometer and gyroscope data, these movements' data tend to resemble the characteristics of falls more closely in terms of accelerometer and gyroscope features compared to the common everyday actions, containing up and down, \ie, putting down and picking up a phone quickly, toward down, \ie, swinging a phone quickly, and \revised{throwing}, \ie, throwing a phone onto a sofa or bed.

The methodology for collecting data on specific actions is structured as follows: 
Firstly, we gather data on falls and common indoor actions, including the first and second categories mentioned above. We then use this data to design and train a machine-learning model, which is subsequently tested in two parts. 

The first part tests actions already known to the model to assess its ability to distinguish these from falls. The second part tests actions not included in the model, such as taking a smartphone out of a pocket to see how this action is categorized. If this action is easily misclassified as a fall, then we specifically collect data on this type of action and retrain the model. The newly trained model is then used to retest this action to see if it can be distinctly separated from falls in the updated model. 

The overall strategy involves training the model and then testing it against any action that might occur in daily life while carrying a smartphone, to identify actions that are prone to being misjudged as falls. Once a misjudged action is identified, we collect data solely on that action, label it separately, and retrain the model.

\begin{figure}[t]
    \centering
    \includegraphics[width=3in]{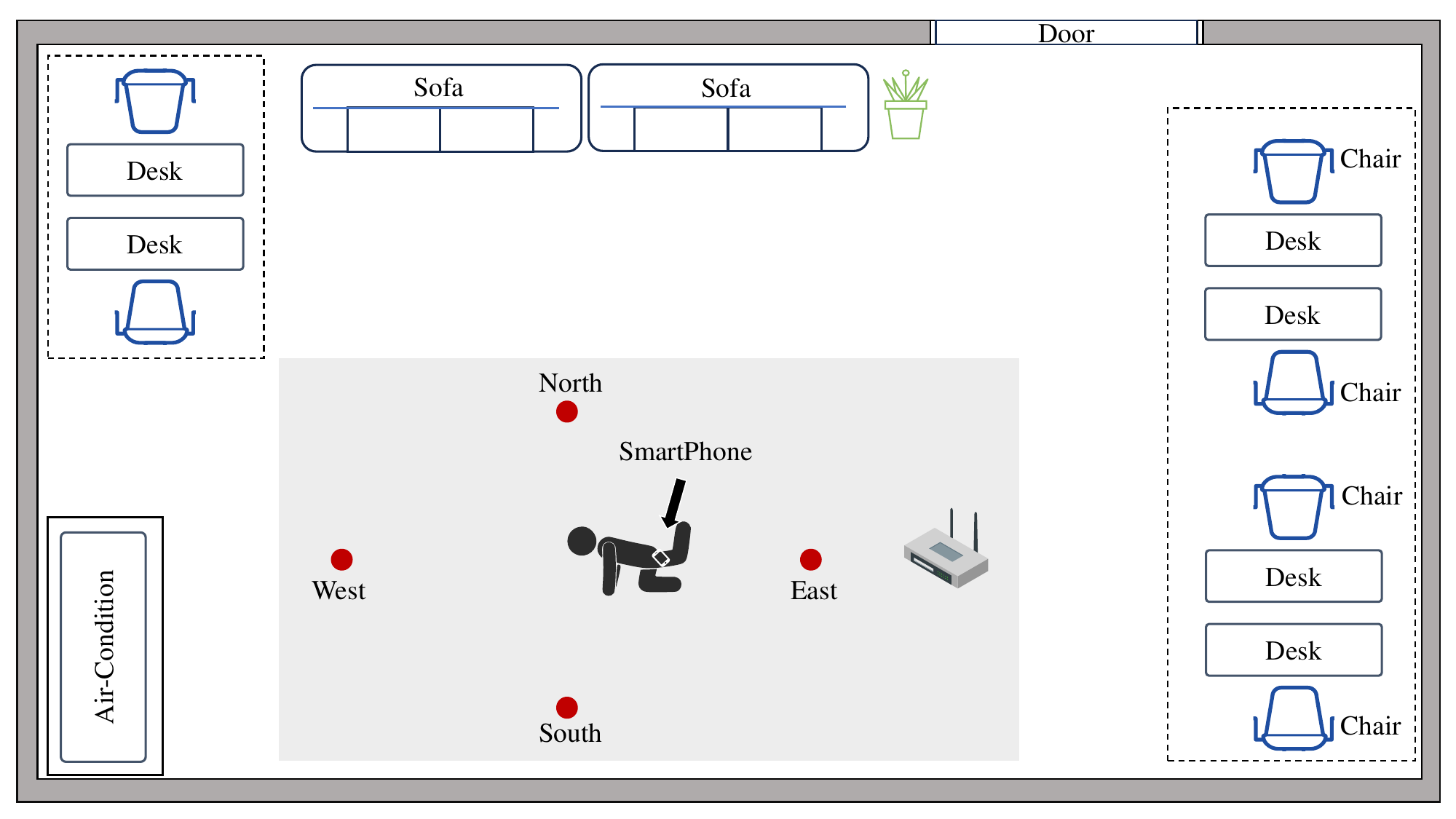}
    \caption{The environment of fall detection using IMU and Wi-Fi CSI built in the smartphones.}
    \label{fig:sys_environment}
\end{figure}
\subsection{Stage I}
The first stage of a fall is defined as a combination of the descent phase, the impact phase, and the onset of the stationary phase, which constitutes the main body of the entire falling process and is characterized by the greatest range of motion. 
Compared to everyday activities such as walking, sitting down, standing up, and squatting, the stage exhibits significant differences in the conditions of acceleration and body rotation during movement.

During the data collection phase, we conduct experiments using a mobile phone in the scenario shown in Fig.~\ref{fig:sys_environment}. The sampling rate of the accelerometer and gyroscope is set to 10~Hz.
To ensure data diversity and experiment accuracy, the experiments are carried out in an environment that both simulates daily life and considers various possible fall situations. 

With an OPPO smartphone carried in their pocket, our participants are asked to simulate various fall behaviors, including slips, trips, and fainting in the four cardinal directions, \ie east, south, west, and north, represented by red dots in Fig.~\ref{fig:sys_environment}. This multi-directional simulation method ensures that the dataset covers all possible fall scenarios, thereby enhancing the reliability of the system in practical applications. 

\begin{figure}[t]
    \centering
    \includegraphics[width=1.25in]{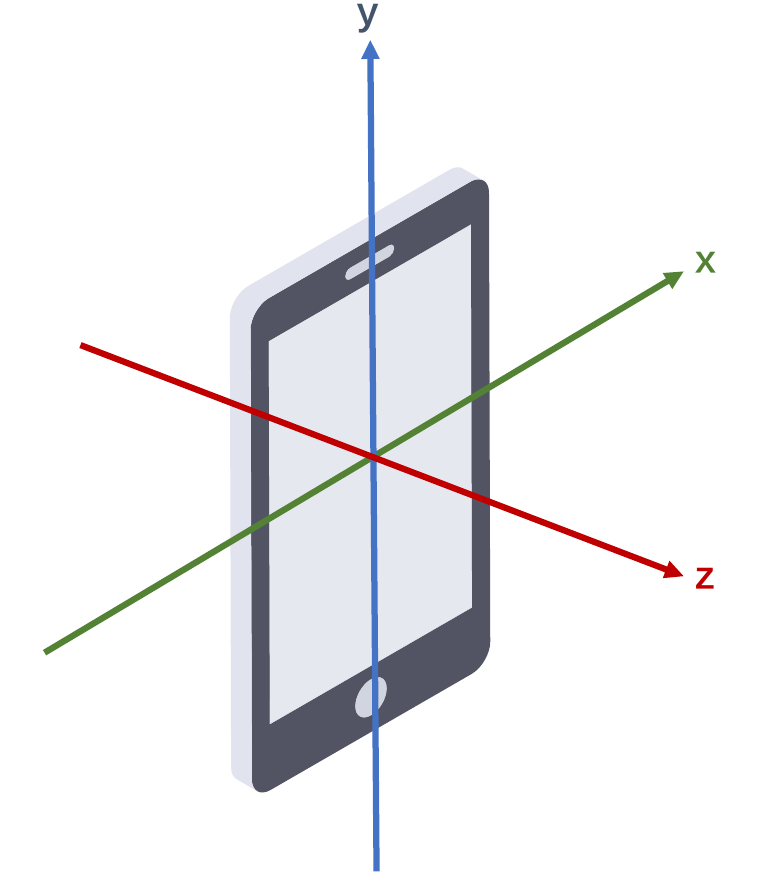}
    \caption{Orientation of built-in sensors of smartphones.}
    \label{fig:sys_coord}
\end{figure}

As illustrated in Fig.~\ref{fig:sys_coord}, when a smartphone is placed in different orientations, the accelerations recorded along the $x$, $y$, and $z$ axes vary due to the presence of inherent gravity. In reality, the position and orientation of a smartphone carried in a pocket differ each time, resulting in varied raw acceleration data even for the same movement. Consequently, it is necessary to remove the influence of gravity on the accelerometer to ascertain the actual acceleration values along the $x$, $y$, and $z$ axes. Subsequently, the acceleration values from these three axes are vectorially combined to derive the acceleration value corresponding to the actual direction of motion.

To remove the gravity component from the raw accelerometer data, a low-pass filter is employed, \revised{as:}
\begin{equation}
    {g_x}_t = \alpha \times {g_x}_{t-1} + (1-\alpha) \times {acc_x}_t
\end{equation}
\begin{equation}
    {g_y}_t = \alpha \times {g_y}_{t-1} + (1-\alpha) \times {acc_y}_t,  
\end{equation}
\begin{equation}
    {g_z}_t = \alpha \times {g_z}_{t-1} + (1-\alpha) \times {acc_z}_t.
\end{equation}
where ${g_x}_t$ denotes the $x$-axis acceleration data at time $t$ after removing the effects of gravity, while ${g_x}_{t-1}$ refers to the $x$-axis acceleration data from the previous moment after gravity has been removed. ${acc_x}_t$ represents the $x$-axis acceleration data collected by the sensor at time $t$, which includes the influence of gravity. 

\begin{figure}[t]
 \centering
   \subfigure[before stage]{
		\begin{minipage}[t]{0.45\linewidth}
            \centering
            \includegraphics[width=0.96\textwidth]{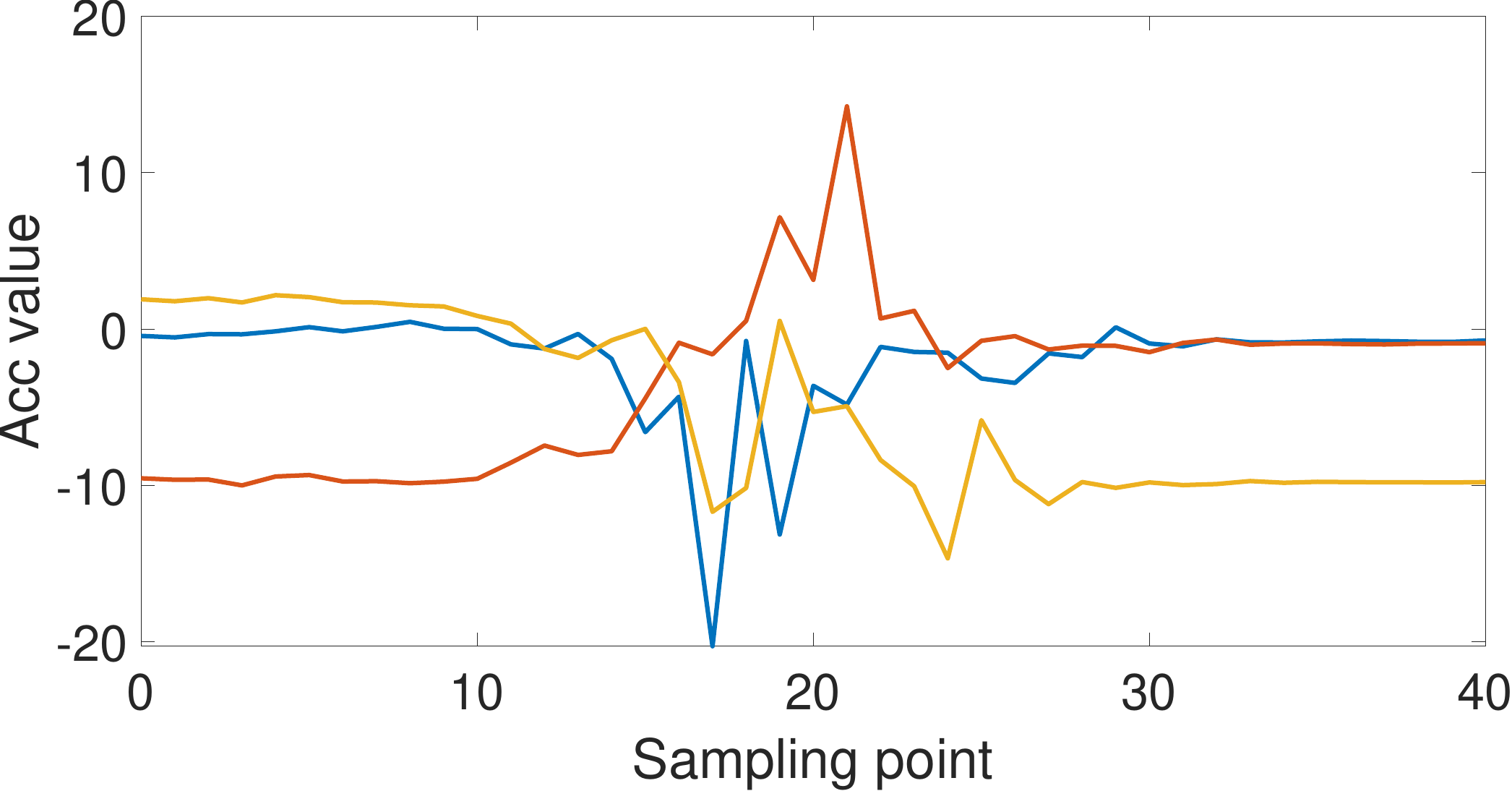}
            \label{fig:acc_have}
            \end{minipage}
		}
    ~
    \subfigure[after stage]{
		\begin{minipage}[t]{0.45\linewidth}
            \centering
            \includegraphics[width=0.96\textwidth]{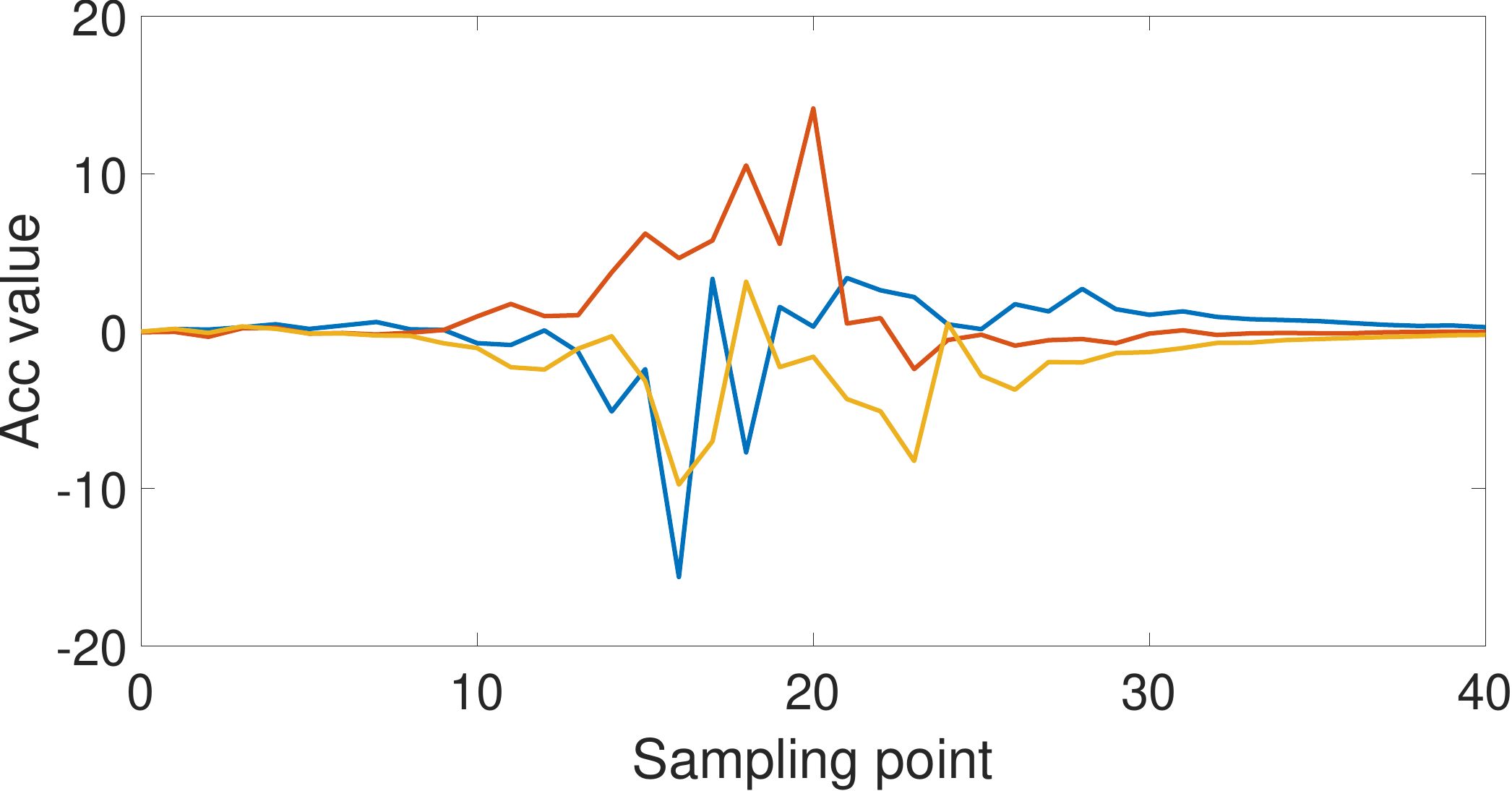}
            \label{fig:acc_nothave}
            \end{minipage}
		}
\caption{Comparison of the acceleration values before and after the removal of \revised{the} gravity.}
\label{fig:gravityremove}
\end{figure}

\begin{figure}[t]
    \centering
    \includegraphics[width=3in]{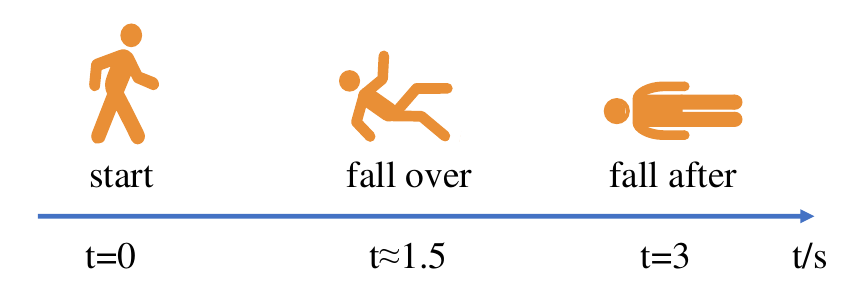}
    \caption{The set-up window of a fall event. To cover the whole duration of a typical fall event, we choose 3~s as a window size.}
    \label{fig:fall_window}
\end{figure}

\begin{figure}[t]
    \centering
    \includegraphics[width=3in]{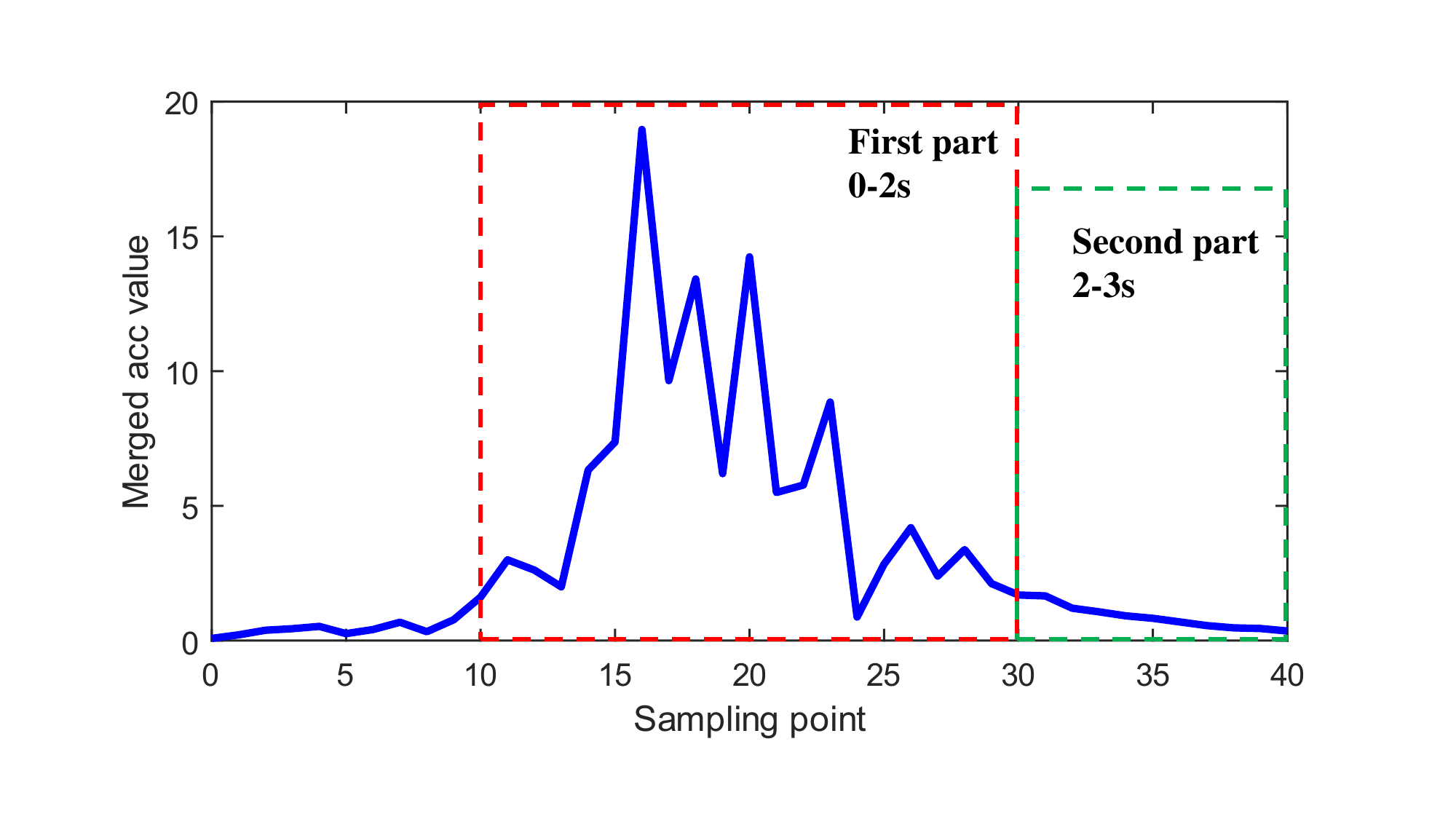}
    \caption{\revised{Values of the acceleration synthesized along three axes, further subdivided into two parts for the initial stage of the fall.}}
    \label{fig:acc_syn}
\end{figure}

\begin{figure}[t]
    \centering
    \includegraphics[width=3in]{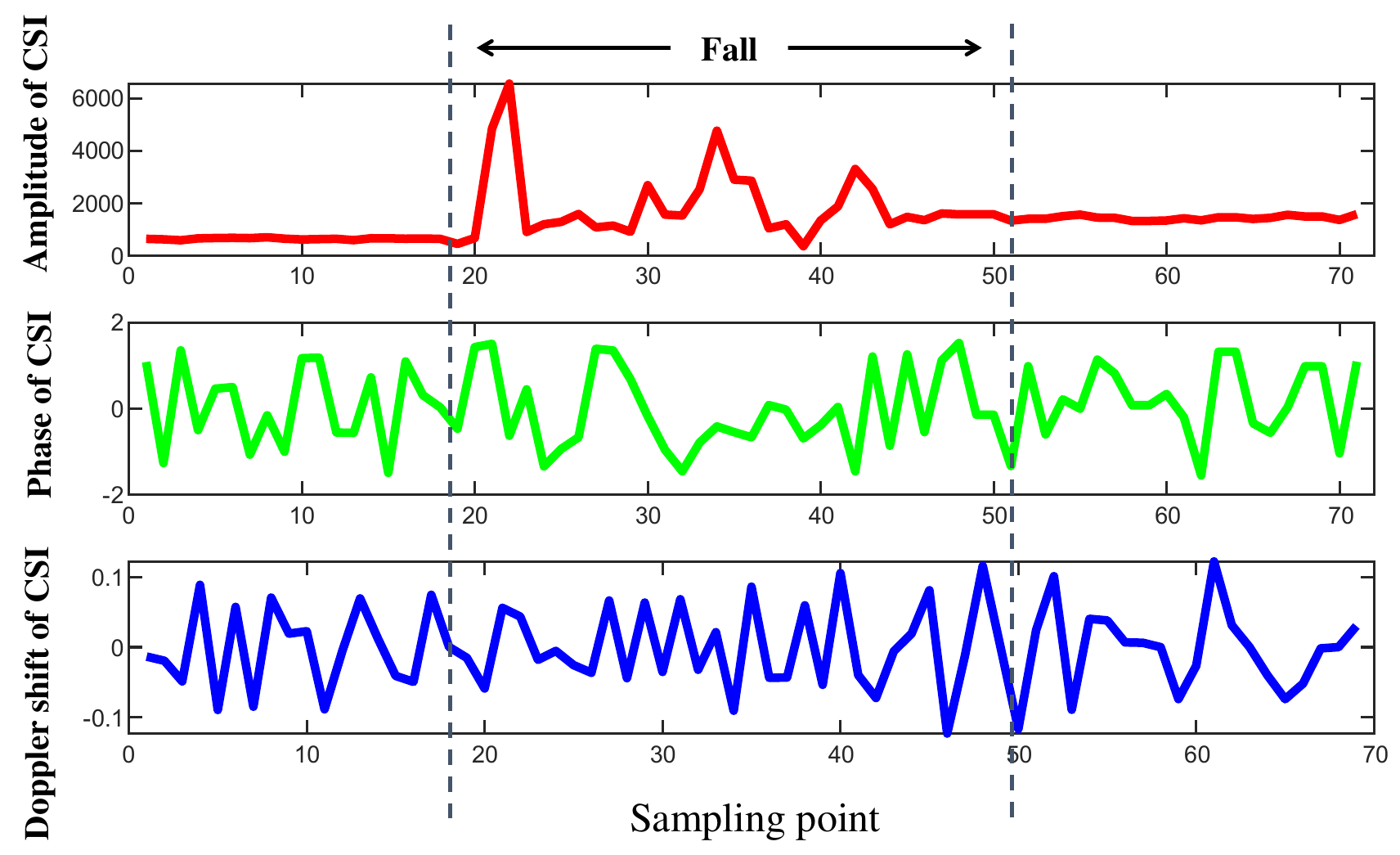}
    \caption{\revised{Phase, amplitude, and Doppler shift of CSI corresponds to a fall event.}}
    \label{fig:csi_p&a&d}
\end{figure}

An exponential weighted moving average filter is used here to smooth the output of the accelerometer, reducing the impact of high-frequency noise and vibrations. The parameter $\alpha$ represents the weighting relationship between the current and past data, where $\alpha$ is a constant between 0 and 1. If $\alpha$ is close to 1, the current data has a greater weight, resulting in a stronger filtering effect. Conversely, if $\alpha$ is closer to 0, the historical data has more weight, and the filtering effect is weaker. For this case, $\alpha$ is set to 0.85.

To compute the linear acceleration, we ascertain the object's true acceleration in space after the removal of gravitational effects:
\begin{equation}
    lacc_{x_t} = acc_x - g_x,
\end{equation}
\begin{equation}
    lacc_{y_t} = acc_y - g_y,
\end{equation}
\begin{equation}
    lacc_{z_t} = acc_z - g_z,
\end{equation}
where $lacc_{x_t}$, $lacc_{y_t}$, and $lacc_{z_t}$ denote the actual acceleration values of a mobile phone moving in the $x$, $y$, and $z$ axes, respectively. The comparison of the acceleration values before and after the removal of gravity is shown in Fig.~\ref{fig:gravityremove}.

Our data indicates that it takes approximately $1.5$ seconds from the initiation of a fall to the impact \revised{stage}. Therefore, we define a fall event as occurring within a $3$-second window, as shown in Fig.~\ref{fig:fall_window}. 

Due to the unknown orientation of smartphone placement on the body during human movement, it is necessary to obtain acceleration data independent of the placement direction. Therefore, by adopting the method of vector composition, the accelerations of these three axes are combined to represent the actual acceleration value of the mobile phone, treated as a point mass, as it moves with the human body. This value reflects the acceleration magnitude in the direction of the smartphone's movement. Fig.~\ref{fig:acc_syn} exhibits the values of the acceleration after synthesis along three axes. 

According to the characteristics of the falling action, the data from the first stage of the fall is further subdivided into two parts, \revised{as illustrated in Fig.~\ref{fig:acc_syn}}. 
The first part broadly encompasses the falling and impact phases, while the second part includes a small portion before the stationary phase. The reason is that during the first phase of the approximately 3-second fall, there is a significant disparity between the accelerometer and gyroscope characteristics between 0-2 seconds and 2-3 seconds. Each of these segments exhibits distinct features. In particular, within 1-2 seconds, the value of the accelerometer rapidly increases from around zero and then quickly drops back to zero. Throughout this process, the average value of the accelerometer is relatively high, and the rate of change is substantial. The maximum value of the accelerometer during these 3 seconds generally appears within the initial 0-2 seconds. The subsequent phase, which involves lying on the ground for about 1 second after the fall, is characterized by minimal changes and average values in both the accelerometer and gyroscope.

\begin{table*}[t]
    \centering
    \caption{Composition of continuously collected CSI data} 
    \label{CSI_data}
    \begin{tabular}{c|c|c|c|c}
        \hline
        \multirow{2}{*}{\diagbox{Sample time}{Single sampling}} & \multicolumn{2}{c|}{Real part} & \multicolumn{2}{c}{Imaginary part}\\
        \cline{2-5}
         & First antenna & Second antenna & First antenna & Second antenna \\
        \hline
        t & 0-53 & 0-53 & 0-53 & 0-53\\
        \hline
    \end{tabular}
\end{table*}

\begin{table}[t]
\centering
\caption{\revised{Parameters of MLP model used for processing IMU data.}}
\begin{tabular}{ccc}
\hline
\textbf{Layer}      & \textbf{Output Shape} & \textbf{Param} \\ \hline
Dense           & (None, 128)           & 2688             \\ \hline
Dropout          & (None, 128)           & 0                \\ \hline
Dense          & (None, 64)            & 8256             \\ \hline
Dropout       & (None, 64)            & 0                \\ \hline
Dense           & (None, 32)            & 2080             \\ \hline
Dense           & (None, 11)            & 363              \\ \hline
\end{tabular}
\label{tab:IMU_model_params}
\end{table}
As mentioned earlier, the whole dataset comprises ten distinct actions, each split into a training set and a testing set in a 7:3 ratio. Each input sample to the model is a one-dimensional array containing 20 features, \ie, maximum, mean, median, kurtosis, and variance of the first and second part of accelerometer and gyroscope data. 

The input is first fed into a fully connected layer, where each feature is connected to 128 neurons and processed through a ReLU activation function, resulting in a one-dimensional array of 128 values. 
This is followed by the first Dropout layer, which randomly sets the weights of input neurons to zero, with a dropout rate of 10\% to prevent overfitting. 
The process continues with a second fully connected layer, taking the output of the previous layer and processing it through a ReLU activation function to produce an array of 64 values. 
A second dropout layer follows, randomly dropping 5\% of the neurons to further prevent overfitting. 
The flow then moves to a third fully connected layer, which takes the output from the previous layer, a 64-value one-dimensional array, connects it to 32 neurons, processes it through a ReLU activation function, and outputs a 32-value one-dimensional array. 
Finally, the output layer takes this 32-value array, connects it to 11 neurons, and activates them with a softmax function to produce an 11-value one-dimensional array, which represents the probabilities of the different actions. 
\revised{The usage of a cross-entropy loss function with the softmax function as the activation for neural network outputs transforms the raw numerical output of the model into a probability distribution, ensuring outputs for each category are between [0,1] and their total equals 1. Thus, the outputs can be interpreted as category probabilities. The model learns to map inputs to category probabilities and selects the one with the highest probability as the prediction. By applying the backpropagation algorithm, the model adjusts its weights to minimize the loss, improving classification accuracy for each category.}
The parameters of IMU model is indicated in Table~\ref{tab:IMU_model_params}.

\begin{figure}[t]
    \centering
    \includegraphics[width=3in]{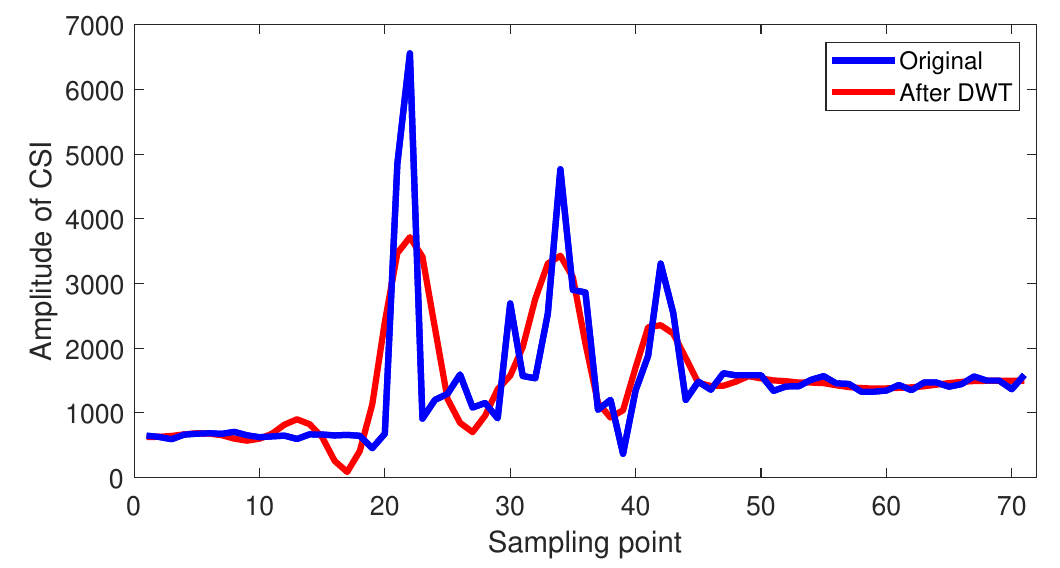}
    \caption{CSI after DWT.}
    \label{fig:csi_dwt}
\end{figure}
\subsection{Stage II}
The primary purpose of this stage is to check whether or not the user still has the ability to move in the stationary phase, using the Wi-Fi CSI data. Based on the configuration of the Wi-Fi CSI tool, the beacon's transmission frequency is typically set at 10~Hz, consequently, the sampling rate for the CSI data captured from beacon frames is also 10~Hz. 

CSI data is collected from two receiving antennas on our mobile device, with each antenna gathering information from $53$ subcarriers. The data for each subcarrier is further divided into the real and imaginary components of the CSI. 

Thus, the corresponding dimensions of the raw CSI data are $30 \times212$. The $30$ rows indicate that, due to the 10~Hz sampling rate, a total of $30$ CSI values are captured over 3 seconds, and the $212$ columns correspond to the real and imaginary parts of the CSI data captured, summing up to $212$ components from $106$ subcarriers across two antennas.

From the collected CSI data, both the real and imaginary parts can be used to calculate the amplitude and phase of the CSI data. CSI provides granular information about the signal during transmission, including multipath effects and phase changes. 
Multipath effects refer to the phenomenon where Wi-Fi signals reach the receiver through multiple paths due to different reflections, refractions, and diffractions encountered along the propagation path. These signals from various paths cause phase superposition at the receiver, leading to changes in the signal phase, \revised{which can be described as follow:} 
\revised{
\begin{equation}
    s(t) = \sum_{i=1}^{N} A_i e^{j(\omega t + \phi_i)}
\end{equation}
where $s(t)$ is the received signal at time $t$, $N$ is the number of paths the signal has taken, $A_i$ is the amplitude of the signal from the $i_{th}$ path, $\omega$ is the angular frequency of the signal, $\phi_i$ is the phase shift introduced in the signal from the $i_{th}$ path due to path length differences, $j$ is the imaginary unit.
}
In CSI data, multipath effects manifest as rapid phase changes, making it challenging to extract phase information from CSI. This difficulty makes it hard to correlate CSI phase data with the relative stillness and movement actions in the second phase of fall detection. 
\revised{
What's more, in wireless communication, the Doppler effect manifests as a shift in the phase and amplitude of the received signal, attributable to the relative motion between the transmitter and receiver. 
This phenomenon is quantified as follow:
\begin{equation}
    \Delta f = \frac{v}{c} f_0 \cos(\theta)
\end{equation}
where $\Delta f$ represents the Doppler shift, $v$ denotes the velocity of the observer relative to the source, $c$ is the speed of light, $f_0$ refers to the original transmitted frequency, and $\theta$ is the angle of movement relative to the line of sight. 
In the realm of CSI, these principles facilitate the detection of motion by evaluating phase shifts over time, calculated as:
\begin{equation}
    \Delta \phi = \frac{2\pi \Delta f T_s}{N}
\end{equation}
where $T_s$ denotes the symbol duration and $N$ the number of subcarriers.
}

Hence, we select the amplitude of CSI as the feature. The phase and amplitude waveform \revised{along with the Doppler shift} of CSI is shown in Fig.~\ref{fig:csi_p&a&d}.
The amplitude waveform of the CSI data contains numerous transient and sudden features. Discrete wavelet denoising can effectively remove high-frequency noise while preserving these characteristics, as shown in Fig.~\ref{fig:csi_dwt}, it allows for a clearer analysis of the underlying signal patterns, which are crucial for interpreting the data accurately. 

\begin{table}[t]
\centering
\caption{\revised{Parameters of CNN model used for processing CSI data.}}
\begin{tabular}{ccc}
\hline
\textbf{Layer} & \textbf{Output Shape} & \textbf{Param } \\ \hline
Conv1D & (None, 27, 32) & 10,208 \\ \hline
MaxPooling1D & (None, 13, 32) & 0\\ \hline
Attention Layer & (None, 32) & 32 \\ \hline
Dropout & (None, 32) & 0 \\ \hline
Flatten & (None, 32) & 0 \\ \hline
Dense & (None, 64) & 2,112 \\ \hline
Dense & (None, 2) & 130 \\ \hline
\end{tabular}
\label{tab:CSI_para}
\end{table}
\revised{
The second stage of fall detection is invoked after a fall is detected in the first phase. The purpose is to eliminate instances where throwing a phone is mistakenly identified as a fall. It also distinguishes between falls from which one can recover independently and those from which one cannot. Therefore, following a fall detected by the IMU, the CSI dataset for this phase is futher divided into two categories: prolonged stillness and continued activity. Consequently, our CSI dataset comprises two types of movements: relatively stationary and those with significant amplitude, with 820 instances of the former and 800 of the latter collected.
}

These movements correspond to significantly different rates of change in CSI data, the model employs just a single layer of one-dimensional convolution followed by a fully connected layer, with a softmax activation function in the output layer.
The input data dimension for the one-dimensional convolutional layer is (29, 106), utilizing 32 convolutional kernels of size 3, and the output dimension equals the number of convolutional kernels, which is followed by a one-dimensional max pooling layer with a window size of 2. 
An attention mechanism layer is then added. During its use, the model focuses more on important parts of the input sequence, enhancing its ability to learn key information. 
A dropout layer follows with a 20\% dropout rate to regularize the model and prevent overfitting during training. Then comes a flattened layer, which transforms the multi-dimensional output into a one-dimensional array. 
This is followed by a fully connected layer with 64 neurons and ReLU activation, outputting a dimension of 64. 
Finally, another fully connected layer is added for task classification, using the softmax activation function. 
The overall structure and parameters of the model are outlined in Table~\ref{tab:CSI_para}.
\section{Evaluation}
In order to better compare the classification effect, we introduce three evaluation indexes, \ie, accuracy rate, precision rate and recall rate. For the precision rate, we consider two metrics: TP and FP.

\subsection{Performance Analysis of the IMU Model}
\begin{figure}[t]
    \centering
    \includegraphics[width=2.5in]{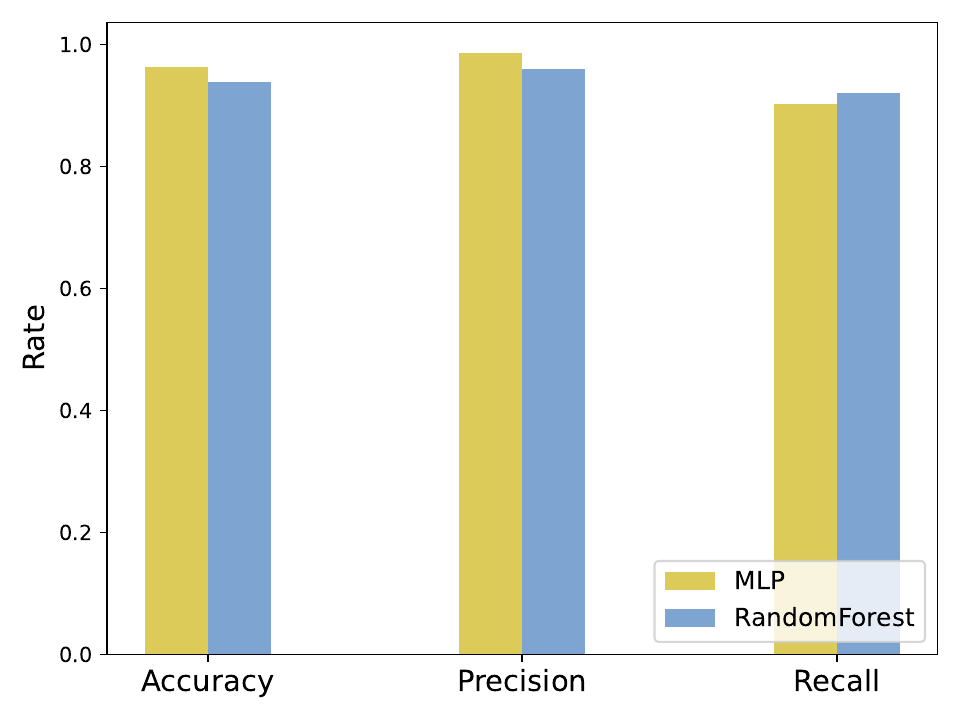}
    \caption{Recognition rate of MLP and Random Forest models.}
    \label{fig:IMU_1}
 \end{figure}

Fig.~\ref{fig:IMU_1} shows the performance of the dataset on MLP and random forest models.

\begin{figure}[t]
    \centering
    \includegraphics[width=2.5in]{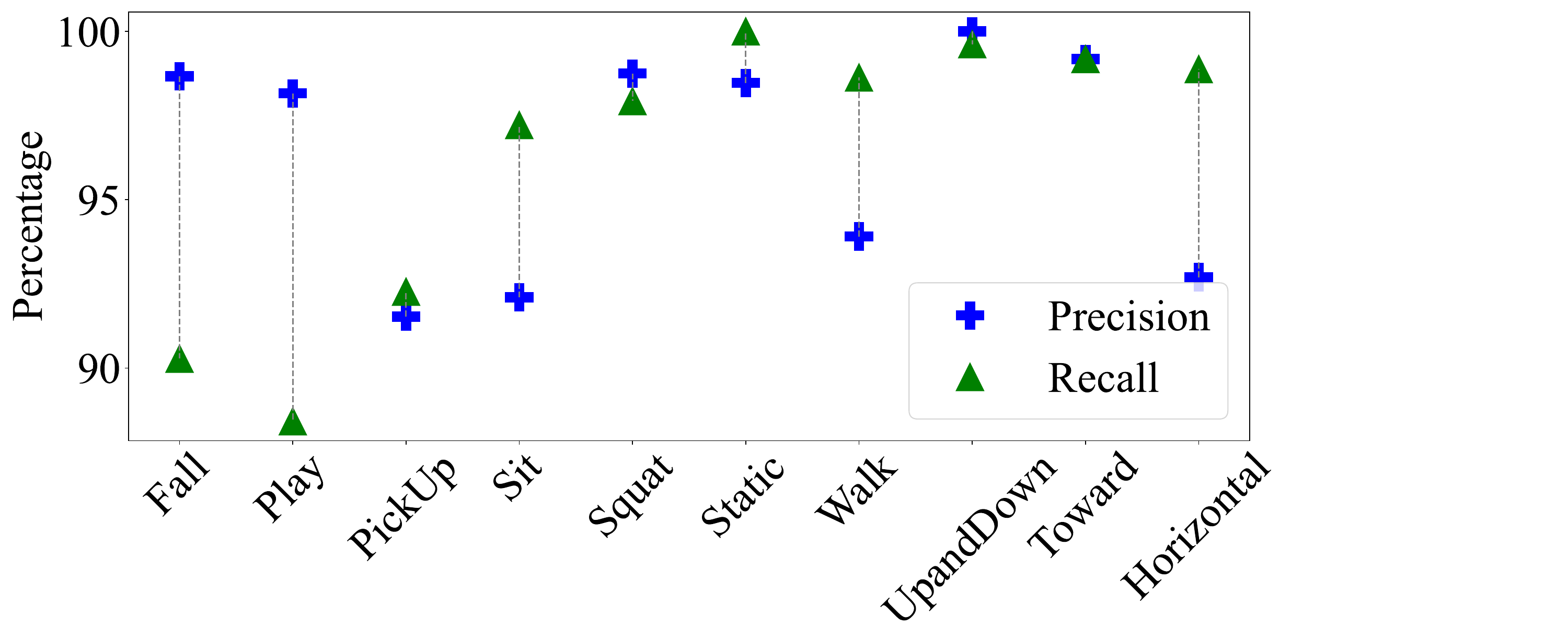}
    \caption{\revised{Precision and recall of MLP model on actual data collected on ten different types of actions.}}
    \label{fig:MLP_actual}
\end{figure}

 \begin{figure}[t]
 \centering
   \subfigure[MLP]{
		\begin{minipage}[t]{0.85\linewidth}
            \centering
            \includegraphics[width=0.98\textwidth]{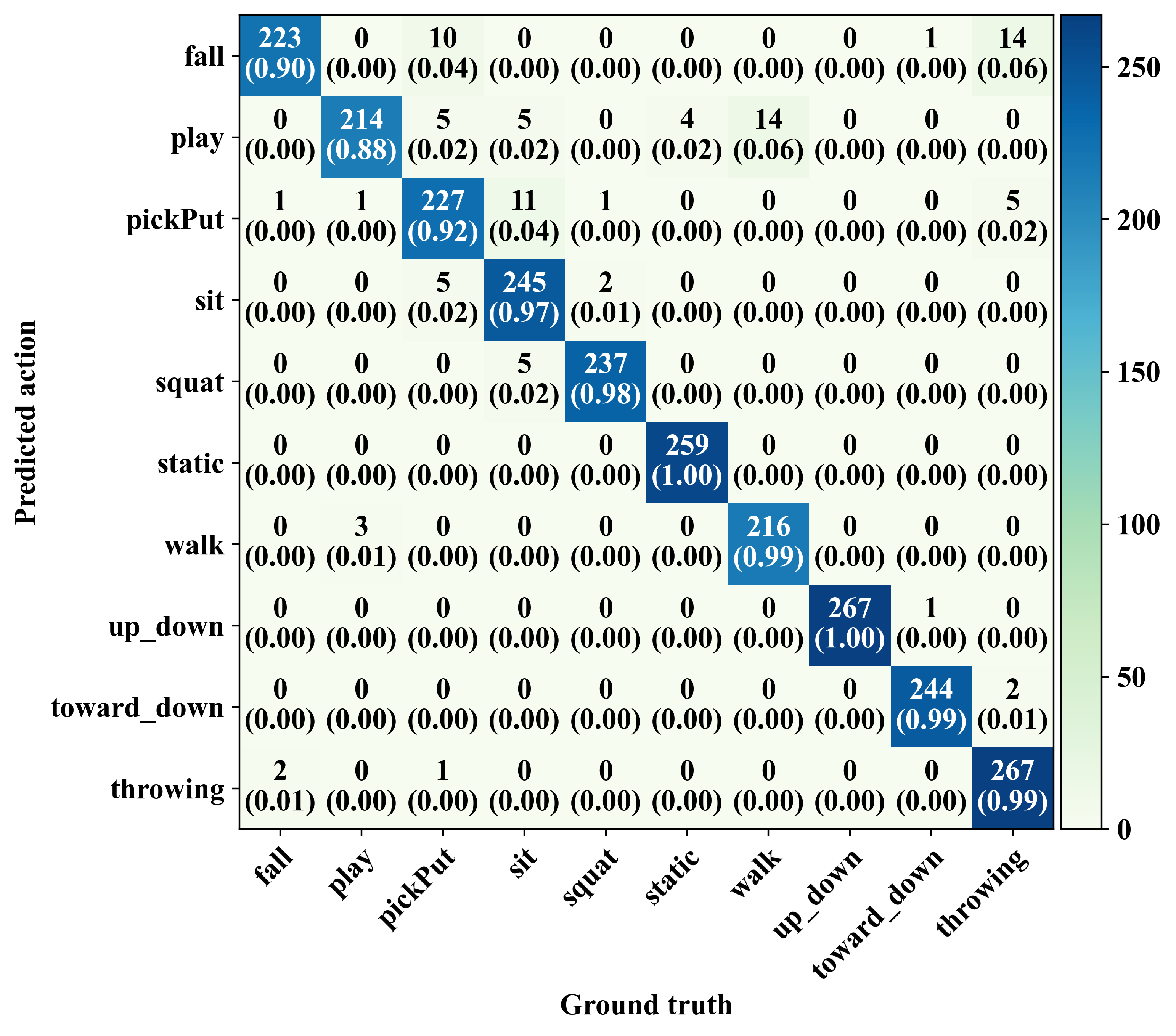}
            \label{fig:confusion_mlp}
            \end{minipage}
		}
        \\
    \subfigure[Random Forest]{
		\begin{minipage}[t]{0.85\linewidth}
            \centering
            \includegraphics[width=0.98\textwidth]{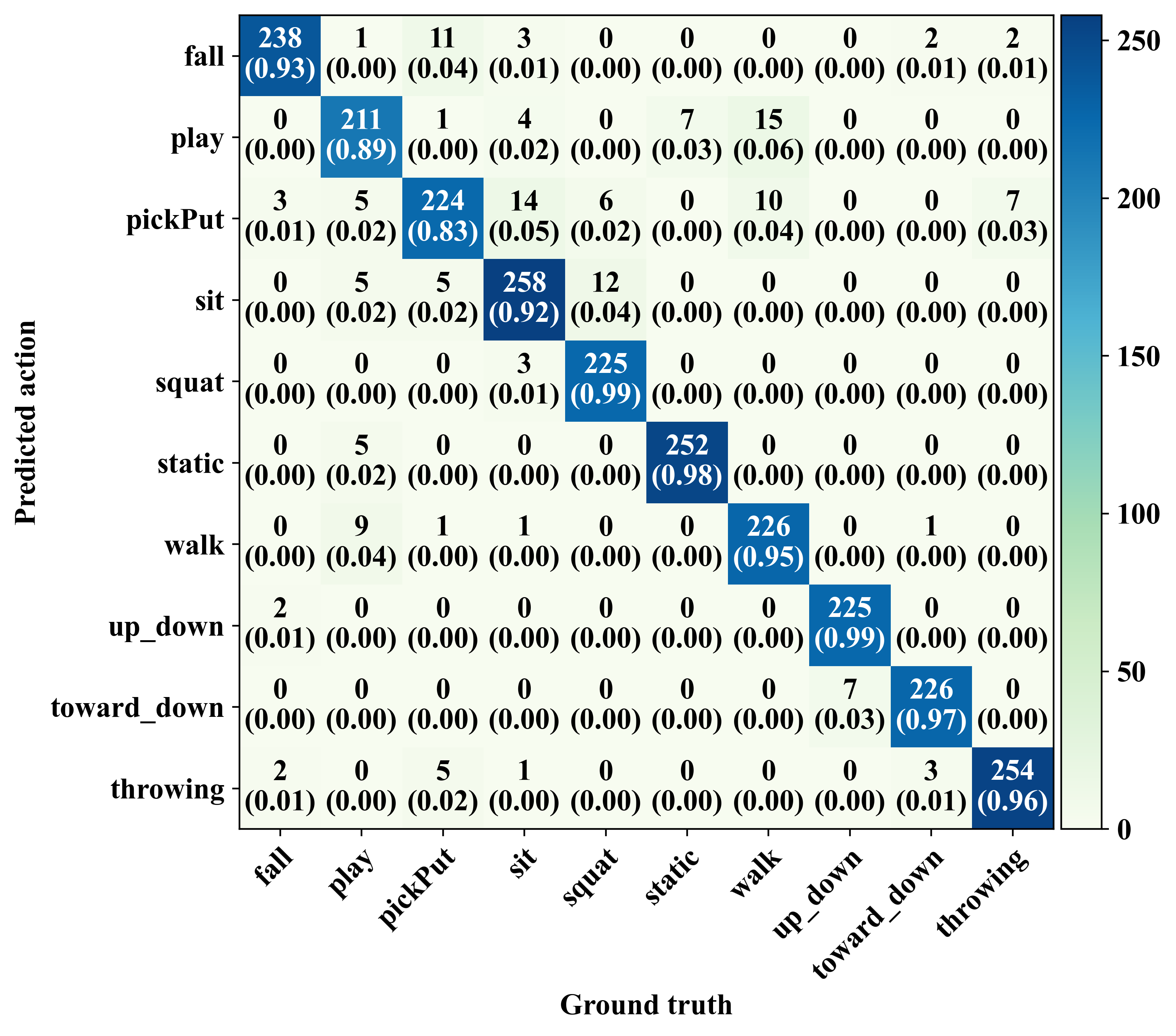}
            \label{fig:confusion_rf}
            \end{minipage}
		}
\caption{Confusion metrics of MLP and RF models on the test set.}
\label{fig:IMU_performance}
\end{figure}
The MLP model built with TensorFlow and the \revised{random forest} model using the {\it sklearn} library on the test set is quite similar across various metrics. The overall accuracy rates for all actions reached 96.31\% and 93.89\%, respectively. To facilitate deployment on Android applications, the MLP was ultimately selected as the classifier to distinguish between falls and other daily activities, achieving a precision of 98.67\% and a recall of 90.28\% on fall actions.

\begin{figure}[t]
    \centering
    \includegraphics[width=3in]{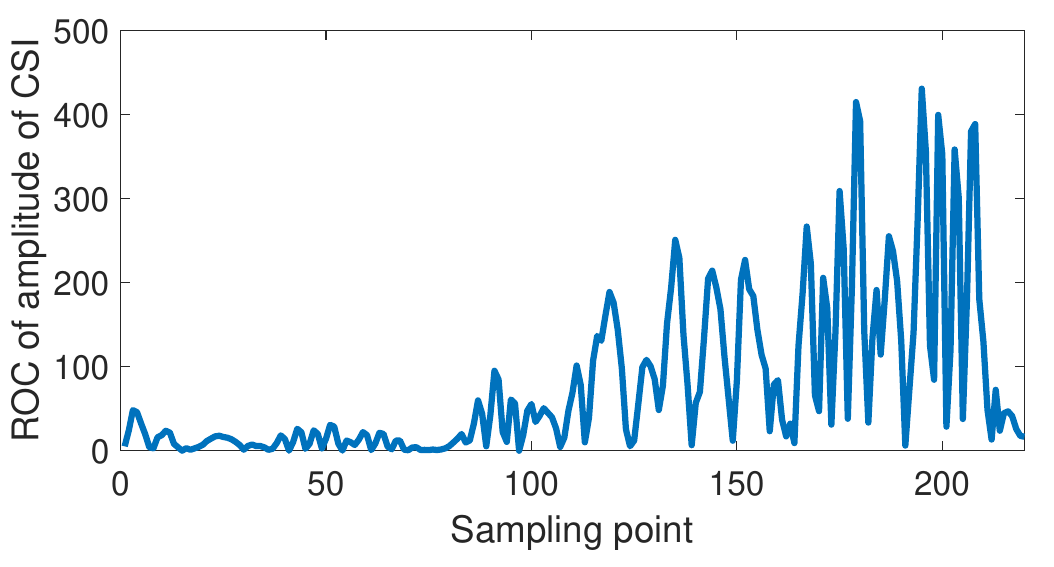}
    \caption{\revised{The ROC of CSI amplitude indoors during periods with or without human activity.}}
    \label{fig:CSI_amplitude_roc}
\end{figure}

\revised{
To evaluate the accuracy of our IMU model, data on specified action types were collected and analyzed. Analysis results, shown in Fig.~\ref{fig:MLP_actual}, indicate that while the model generally distinguishes falls from routine activities effectively, it may not capture all fall events with full accuracy. High precision paired with lower recall for critical actions like ``Fall'' suggests that the model can effectively reduces false positives but may miss true positive events, potentially compromising its effectiveness in real-world applications such as elderly fall detection. Lower recall rates imply that the current model threshold lacks sufficient precision for detecting subtler movements essential to reliable fall detection.
}

Fig.~\ref{fig:IMU_performance} reveals that if a fall action is misclassified, it is generally mistaken for actions such as picking up or putting down a phone, or throwing the phone onto a sofa or bed. Analyzing the actions themselves, these three types of actions exhibit similar changes in accelerometer and gyroscope data. When using these data for the same feature extraction, the extracted features are also quite similar, hence misclassification is inevitable. 

Moreover, following such misclassifications, the subsequent accelerometer readings collected by the phone are very likely to be zero. For instance, after a fall, there might be no further activity, resulting in zero accelerometer data. Similarly, when a phone is quickly put down on a table, it likely remains stationary on the table afterward, also resulting in zero accelerometer readings. The same applies when a phone is thrown onto a bed or sofa, the phone generally remains motionless thereafter. Therefore, accelerometer and gyroscope data become ineffective for further analysis under these conditions. 

Consequently, subsequent judgments have introduced CSI for wireless sensing, \ie, using WiFi signals to detect whether there is any movement in the environment, rather than relying solely on the phone's built-in sensors. \revised{This serves dual purposes. Primarily, it enhances the discrimination of erroneously perceived actions that mimic falls, such as the act of throwing a mobile phone. Typically, following the action of throwing a mobile device onto a surface such as a bed or sofa, an individual is unlikely to remain stationary. Thus, subsequent movements can induce significant disturbances in the spatial channel state, thereby causing notable fluctuations in the collected CSI data.} \revised{Conversely, in instances where a fall results in an inability to self-rescue, such as fainting, there is minimal subsequent movement, which results in a stable channel state, leading to minimal variation in the CSI data. This stability is crucial in further reducing the probability of misjudgment. Fig.~\ref{fig:CSI_amplitude_roc} shows the rate of change (ROC) of CSI amplitude indoors during periods with or without human activity, it can be seen that the first half of the period exhibits no activity, as indicated by the stable CSI waveform. In contrast, the second half shows activity, during which the amplitude of the CSI undergoes violent fluctuations.}

\begin{figure}[t]
 \centering
   \subfigure[Training accuracy]{
		\begin{minipage}[t]{0.45\linewidth}
            \centering
            \includegraphics[width=0.98\textwidth]{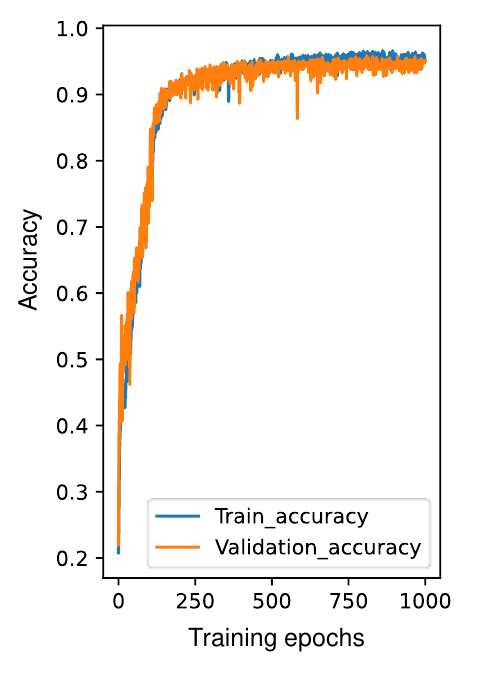}
            \label{fig:accuracy_csi}
            \end{minipage}
		}
    ~
    \subfigure[Loss value]{
		\begin{minipage}[t]{0.45\linewidth}
            \centering
            \includegraphics[width=0.98\textwidth]{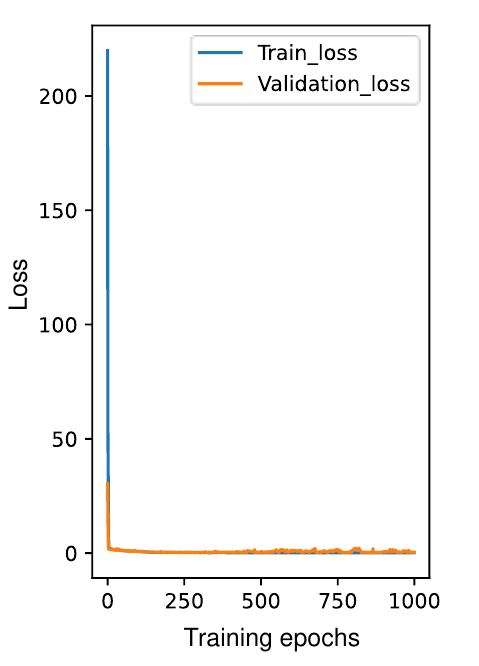}
            \label{fig:loss_csi}
            \end{minipage}
		}
\caption{The accuracy and loss metrics of CSI model.}
\label{fig:CSI_performance}
\end{figure}

\revised{Secondly, the observation of changes in CSI data post-incident enables the determination of whether a fall requiring immediate intervention has occurred. If a fall does occur, subsequent CSI data changes can be observed to determine whether the fall is one that cannot be self-rescued. If it is a non-self-rescuable fall, the subsequent CSI data will remain stable. However, if it is a self-rescuable fall, significant disturbances will appear in the CSI data as the person gets up.}

\revised{
In short, our system distinguishes between actual falls and similar movements by observing subsequent movements. 
True falls, especially those that cannot be self-rescued result in minimal movement, which is reflected in stable CSI data. On the contrary, self-rescuable falls lead to significant disturbances in the CSI data as the person attempts to stand.} 


\subsection{Performance Analysis of the CSI Model}
Fig.~\ref{fig:CSI_performance} illustrates the training and validation performance of our CSI model. The training accuracy swiftly ascends and stabilizes near a value of 1.0, indicative of the model's robust ability to classify training data correctly. Concurrently, the validation accuracy mirrors this trend, suggesting a consistent and generalizable performance across unseen data, without significant overfitting, as the curves for both training and validation accuracy closely align and stabilize at high values. 
Loss metrics for both training and validation initially display a steep decline, leveling off to indicate minimal error as the model converges. The closeness of the training and validation loss curves further substantiates the absence of overfitting, reinforcing the model's effective generalization capabilities. 

Exhibiting outstanding convergence behavior, the CSI model consistently achieves high accuracy and maintains low loss across both training and validation sets. This performance underscores its effectiveness and potential for real-world applications, with minimal risk of overfitting.

\begin{figure}[t]
    \centering
    \includegraphics[width=2in]{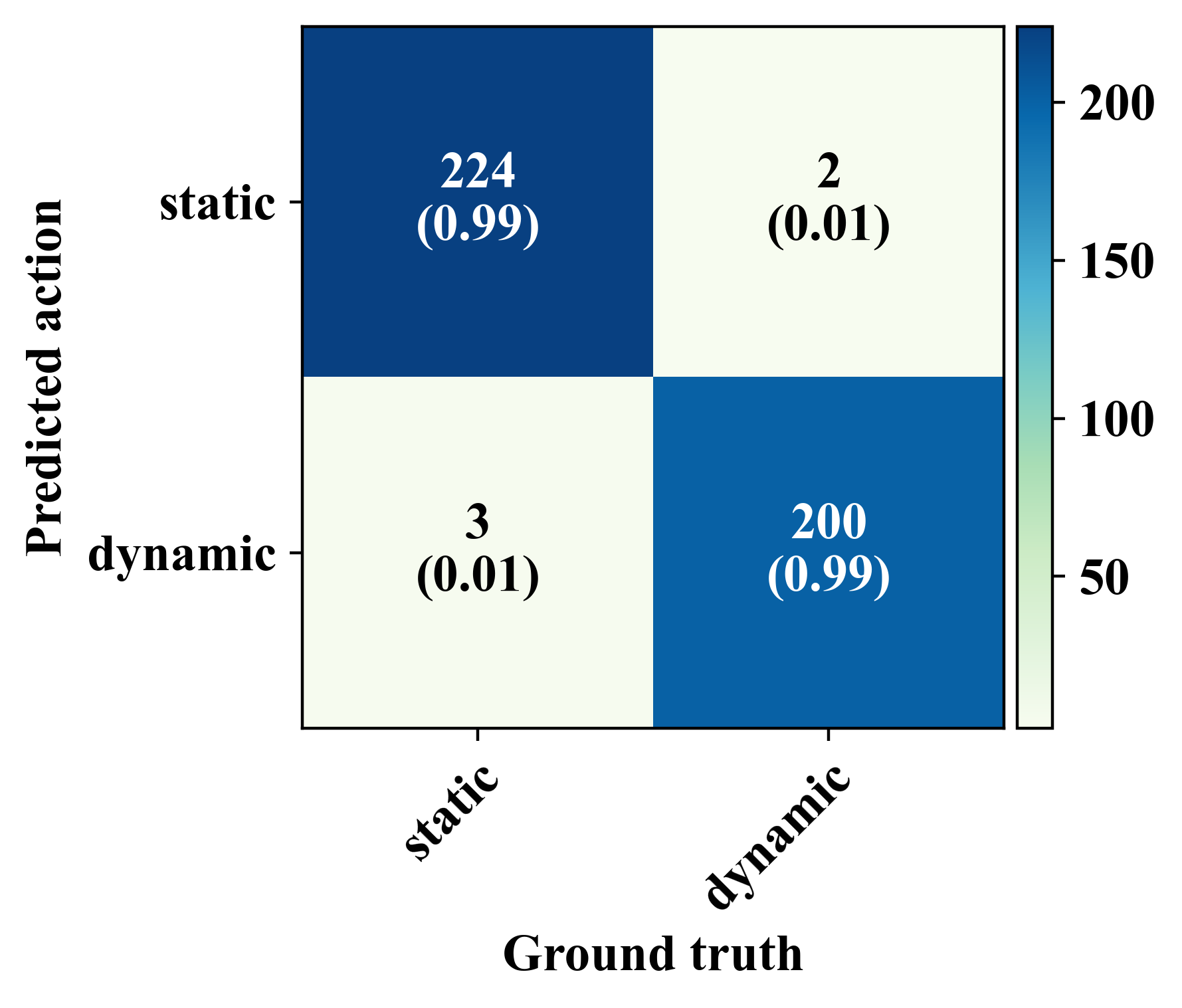}
    \caption{Confusion metrics of CSI models.}
    \label{fig:csi_confusion}
\end{figure}

Fig.~\ref{fig:csi_confusion} shows that in a stationary state, the values of each subcarrier may not be fixed, but as long as the position of the phone remains unchanged and there is no human activity in the space, the variation rate of each subcarrier remains very low, typically below $50$. However, human movement within the space can cause changes in the amplitude of the CSI data, leading to a variation rate much higher than when relatively static. In the test set, these two types of actions achieved a 99\% accuracy rate on the CSI model designed.

\begin{figure}[t]
    \centering
    \includegraphics[width=3in]{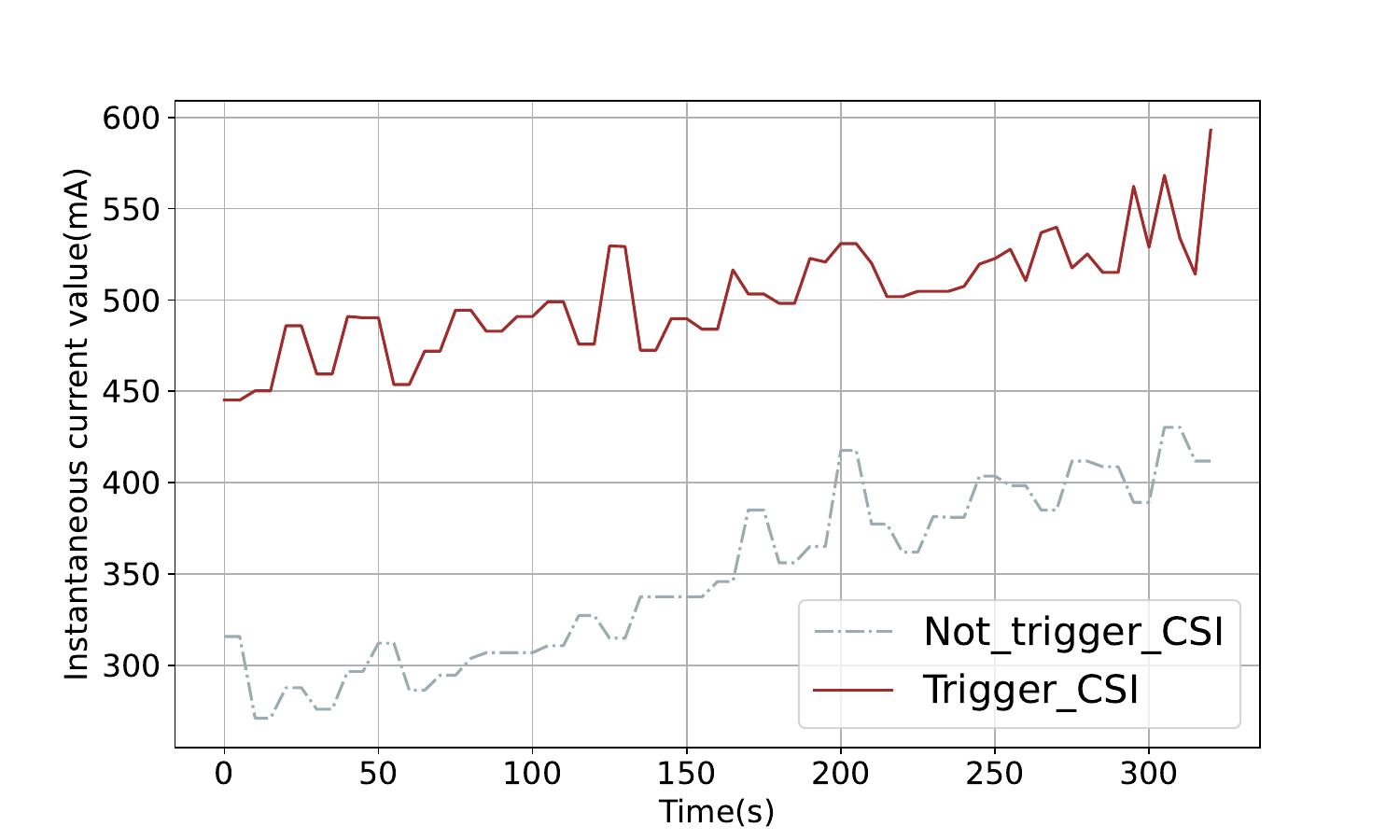} 
    \caption{\revised{Power consumption comparison of whether to continuously trigger the CSI model.}}
    \label{fig:csi_consumption}
\end{figure}

\revised{
Our system achieves an accuracy that is 1.28\% greater than that of the ANN reported in~\cite{cheffena2015fall} when utilizing spectrogram features. It also surpasses the fall detection system proposed in~\cite{lin2020fall} by 3.64\%, and is 3.14\% more accurate than the research~\cite{ferreira2022wearable} conducted by Felipe~\etal
}

\revised{
Furthermore, Our system not only enhances accuracy but also resolves key issues such as reducing false positives and maintaining user compliance without extra hardware, streamlining deployment and increasing user adherence~\cite{li2012microphone,mobsite2024privacy}.
}

Above all, it is evident that our CSI model effectively distinguishes between static and dynamic states. The experimental results demonstrate that the IMU model can account for the majority of fall incidents. Consequently, by augmenting the IMU model with the CSI model, it is possible to provide warnings and call for assistance by assessing whether an individual retains the ability to move after a fall.

\revised{
It is pertinent to mention that the exclusive use of the CSI model is not advisable primarily due to substantially higher power consumption. Fig.~\ref{fig:csi_consumption} presents a comparison of power consumption when the CSI model is activated continuously, illustrating the inefficiencies of its standalone use. The graph shows the trigger condition for CSI activation (depicted in red), maintaining consistently higher values over time compared to scenarios where the CSI model is not triggered (dotted grey line). This clear difference in power usage highlights the necessity of integrating more power-efficient models such as the IMU to enhance overall system efficiency.
}

\begin{figure}[t]
    \centering
    \includegraphics[width=3in]{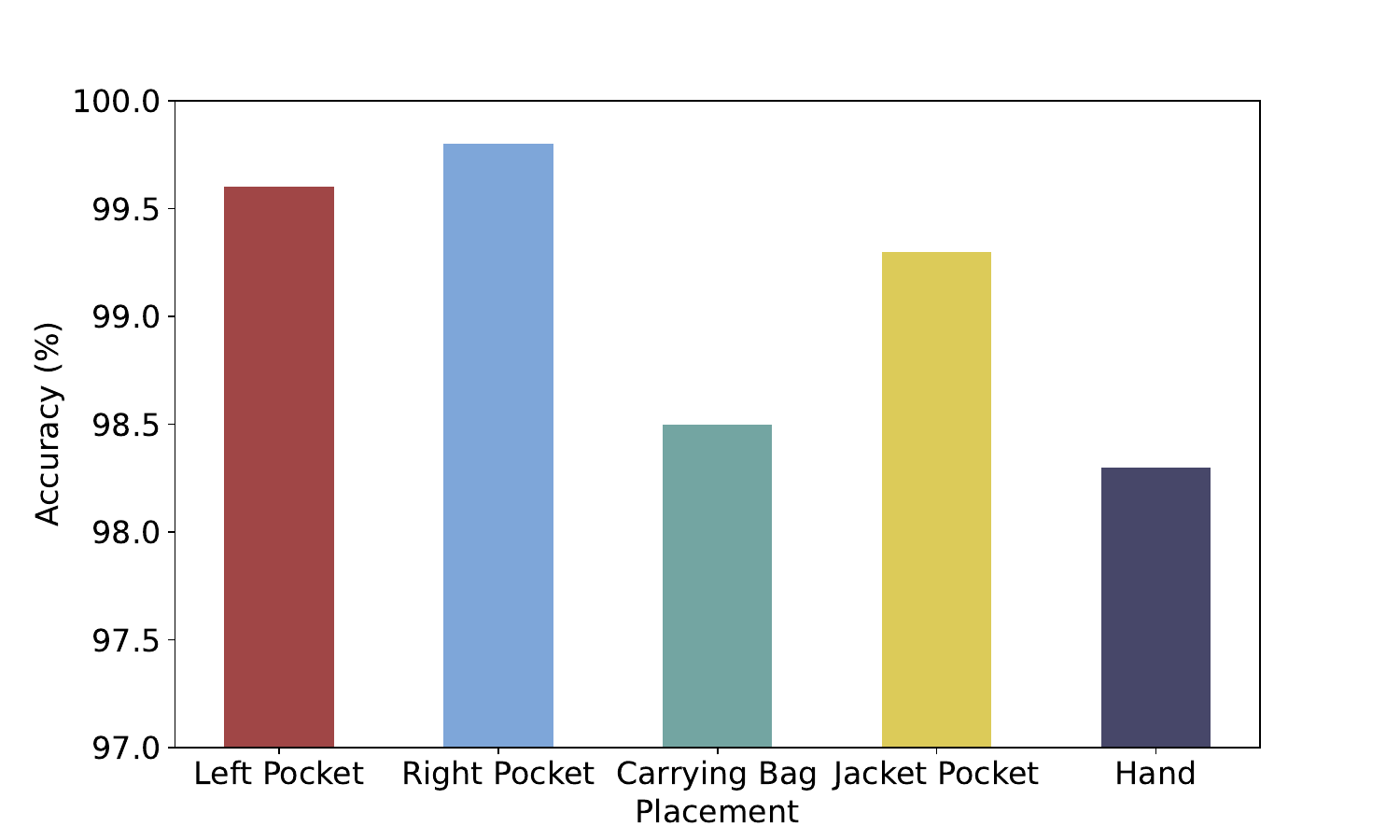} 
    \caption{\revised{The accuracy of the system in detecting falling movements at five different locations: left pocket, right pocket, carring bag, jacket pocket, hand.}}
    \label{fig:csi_placement}
\end{figure}

\revised{
Considering the potential impact of varying phone placements on model accuracy, as shown in Fig.~\ref{fig:csi_placement}, we conducted verification experiments to ascertain that environmental factors minimally affect the overall system detection capabilities. This is due to the reliance on accelerometers and gyroscopes within the device to complete its functions. Furthermore, given that the primary objective of CSI is to detect motion changes in the environment, the presence of such changes ensures that our system will consistently generate a result. On average, our system for fall detection achieved an accuracy rate of 99.1\% across five distinct directions.
}

\subsection{Deployment and Testing of the APP}

The fall detection APP is based on the Android 13 system and utilizes version 33 of the Android SDK to provide application interfaces and libraries for development. The source code is set to be compatible with Java 8, meaning it is compiled according to the syntax rules of Java 8. The Gradle build tool is used to manage the project's building, dependencies, version control, resource management, and other tasks.
The IMU and CSI models deployed on the Android system have been converted into the TensorFlow lite format.

The fall detection process is monitored with a step size of 0.1 seconds. During the entire sliding window process, approximately 5-10 fall detections are likely to occur. However, for other actions similar to falling, such as picking up or placing a phone, it is rare for the detection to register 5-10 falls throughout the sliding detection process. To enhance the sensitivity of fall detection and reduce the probability of misclassifying other actions as falls, the real-time detection results are stored in a buffer, which has a capacity of 20. Once the buffer is full, the oldest result is discarded each time a new result is entered. The system then continuously monitors the number of falls among these 20 results in the buffer. In practical tests, setting the detection criteria to identify a fall if 3 out of 20 results indicate a fall provided the best overall performance.

\begin{figure*}[t]
\centering
\begin{tabular}{cccc}
\includegraphics[width=0.2\textwidth]{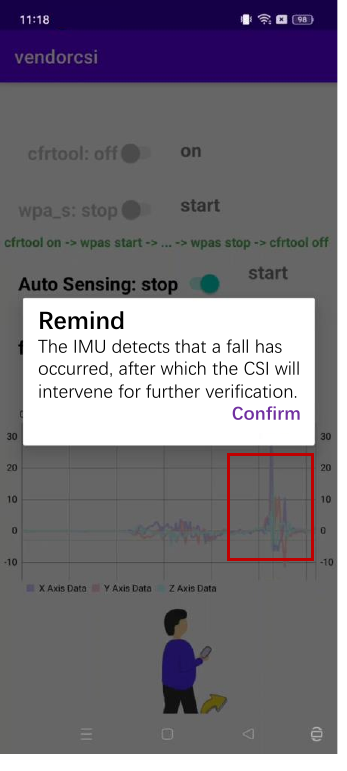} &
\includegraphics[width=0.2\textwidth]{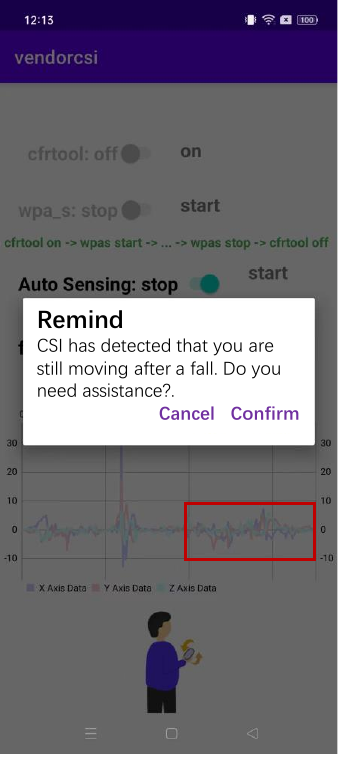} &
\includegraphics[width=0.2\textwidth]{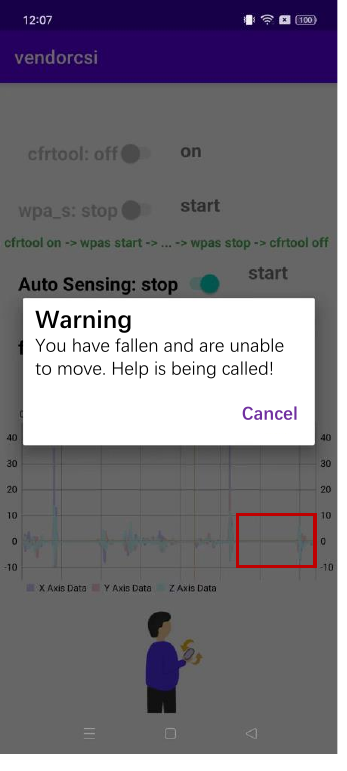} &
\includegraphics[width=0.2\textwidth]{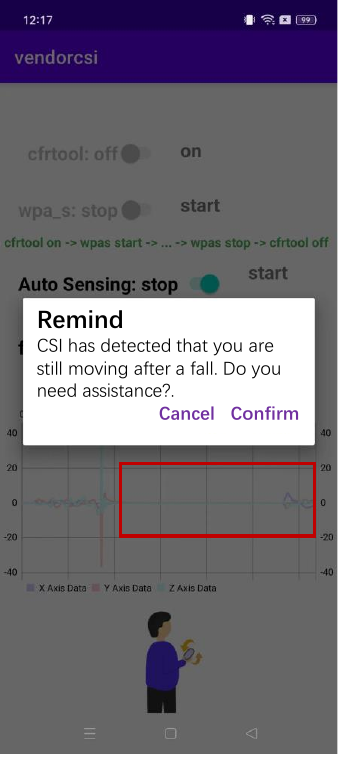} \\
(a) IMU detected as a fall & (b) self-rescuable fall & (c) non-self-rescuable fall & (d) misjudgment
\end{tabular}
\caption{Results of the fall detection by the app.}
\label{fig:res_app}
\end{figure*}

Fig.~\ref{fig:res_app}(a) demonstrates the prompt for the IMU model to call the CSI model after detecting a fall. Fig.~\ref{fig:res_app}(b) depicts a real fall after which the person manages to get up, classified as a fall followed by an activity, indicating a self-rescuable fall. Comparing Fig.~\ref{fig:res_app}(c) and Fig.\ref{fig:res_app}(d), in Fig.~\ref{fig:res_app}(c) there is little to no movement after the fall, which is confirmed by the accelerometer data visualization indicating no movement post-fall, classifying it as a non-self-rescuable fall, which is the type of fall behavior our system aims to detect. 

In contrast, Fig.~\ref{fig:res_app}(d) results from a misclassification caused by throwing the phone onto a sofa, whose accelerometer data shows resulted in no subsequent movement of the phone. However, human activity still occurs in the space, causing significant fluctuations in subsequent CSI data, leading to the determination that this fall detection was a misjudgment of another action. This helps address the first stage IMU model's shortcoming of insufficient accuracy in distinguishing falls from fall-like actions.
\section{Conclusion}
This \revised{study} presents an innovative real-time fall detection system, utilizing built-in hardware, \ie, accelerometers and gyroscope sensors, and combines them with commercial WiFi on Android smartphones. We review the research background and existing technologies addressing the challenges within fall detection systems, drawing on studies from both domestic and international contexts. A sensor-based method is explored in detail, involving the collection of data from accelerometers and gyroscopes during movement. This approach proposes a nuanced process for handling raw data of fall events tailored to different stages of a fall. By doing so, it enhances the feature extraction from the fall process and introduces a sensor orientation-independent fall detection model.

\revised{
One of the intrinsic limitations of the smartphone-based fall detection system presented in this study is its dependency on the device’s battery life. Since continuous monitoring is essential for ensuring the safety of individuals, particularly the elderly, battery depletion could compromise the system's effectiveness. To address this concern, we propose the integration of an automated low-battery alert within the application. This feature would actively monitor the device’s battery status and notify the user with an audible alarm or visual prompt when the charge level falls below a predetermined threshold, urging them to recharge the device promptly.
}

The analysis identifies actions that are commonly misclassified by sensor-based fall detection systems and their drawbacks. We further propose the use of CSI as a supplementary criterion in the model, effectively resolving the issue of the inability to collect sensor data when the phone is stationary, thus enhancing the reliability and completeness of the entire fall detection system. To sum up, this article describes the deployment of IMU and CSI models within an Android-based fall detection App, detailing the system's capability to collect data, process it, and utilize both models in real-time fall detection.


\section{Discussion and Future Work}
Leveraging the distinctive capabilities of sensors and CSI, we develop a novel real-time, mobile-based fall detection system, achieving preliminary results and laying a foundation for further investigation. Despite these advances, the project is constrained by limited theoretical insights and research capacity, presenting numerous avenues for enhancement. 

For instance, the current methodology employs a smartphone positioned in a pocket to gather data pertinent to fall detection, which may not accurately capture the dynamics of a fall as the center of balance is located in the cerebellum. Subsequent studies could investigate more effective devices for monitoring head movements during falls, thereby improving the accuracy of fall detection. 
Additionally, to enhance the model's generalizability, statistical features such as mean, variance, and kurtosis are extracted from the raw data. Nevertheless, these features may not fully encapsulate the complexity of the raw data, prompting a need for refined data processing techniques and feature extraction methodologies to more effectively represent the pertinent information. 
Furthermore, the existing CSI model incorporates data from all subcarriers, resulting in extensive data volumes. Future research could focus on selectively using subcarriers based on real-time channel conditions to streamline data processing, potentially reducing computational demands and energy consumption. 

\revised{What is worth mentioning is that the system’s effectiveness depends on the assumption that users consistently carry their smartphones and maintain a WiFi connection, conditions that may not be reliably met, particularly for elderly individuals who may not habitually carry their phones at home. This limitation is even more pronounced for users with cognitive impairments, such as dementia, who may forget to carry their phones.}

\revised{Recognizing this limitation, future works could involve sensor fusion with alternative sensing modalities. While our primary focus in this study is on exploring the feasibility and accuracy of smartphone-based technology, we acknowledge that the inclusion of complementary sensors could enhance system reliability. Sensor fusion with wearable or ambient sensors, for instance, could alleviate the dependency on consistent smartphone use and provide a more comprehensive monitoring solution. In the future, researchers could pay more attention on the integration of these complementary sensing modalities to create a hybrid fall detection system.}

Addressing these areas could substantially elevate the system's efficacy and efficiency, rendering fall detection more viable and precise for real-world implementation.
\bibliographystyle{IEEETran}
\bibliography{PhoneCSI}

\end{document}